\documentclass[10pt,journal,compsoc]{IEEEtran}
\usepackage[breaklinks=true,colorlinks,bookmarks=false]{hyperref}

\usepackage{xspace}
\usepackage{graphicx}
\usepackage{booktabs}
\usepackage{multirow}
\usepackage{color}
\usepackage{colortbl}
\usepackage{amssymb}
\usepackage{amsmath}
\usepackage{amsthm}
\usepackage{cite}

\newcommand{\xmark}{\ding{55}}

\usepackage[table]{xcolor}
\definecolor{mygray}{gray}{.92}
\definecolor{mycyan}{cmyk}{.3,0,0,0}
\definecolor{LightCyan}{rgb}{0.95,1,1}

\definecolor{orange}{rgb}{1,0.8,0.6}
\definecolor{yellow}{rgb}{1,1,0.6}

\newcommand{\eg}{\textit{e.g.}}

\newcommand{\etal}{\textit{et al.}}

\newcommand{\x}{\mathbf{x}}

\usepackage[norelsize, linesnumbered, ruled, lined, boxed, commentsnumbered]{algorithm2e}
\usepackage{xcolor}
\usepackage{pifont}

\definecolor{mygray}{gray}{.9}

\usepackage{array}
\newcolumntype{x}[1]{>{\centering\arraybackslash\hspace{0pt}}p{#1}}

\ifCLASSINFOpdf
\else
\fi

\begin{document}

\title{Multimodal Image Synthesis and Editing: \\ The Generative AI Era}
% \title{Multimodal Image Synthesis and Editing: \\ A Survey and Taxonomy}

\author{Fangneng~Zhan,~
        Yingchen~Yu,~
        Rongliang~Wu,~
        Jiahui~Zhang,~
        Shijian~Lu$^{\S}$,
        Lingjie Liu \\
        Adam Kortylewski,~
        Christian Theobalt,~
        Eric Xing,~\IEEEmembership{Fellow,~IEEE}
\IEEEcompsocitemizethanks{
\IEEEcompsocthanksitem F. Zhan is with the Max Planck Institute for Informatics, Germany and the S-Lab, Nanyang Technological University, Singapore.
\IEEEcompsocthanksitem Y. Yu, R. Wu, J. Zhang and S. Lu are with the Nanyang Technological University, Singapore. 
\IEEEcompsocthanksitem L. Liu, A. Kortylewski and C. Theobalt are with the Max Planck Institute for Informatics, Germany.
\IEEEcompsocthanksitem E. Xing is with the Carnegie Mellon University, USA and the Mohamed bin Zayed University of Artificial Intelligence, UAE.
\protect 
\IEEEcompsocthanksitem $\S$ denotes corresponding author, E-mail: shijian.lu@ntu.edu.sg.
}
}

\markboth{IEEE TRANSACTIONS ON PATTERN ANALYSIS AND MACHINE INTELLIGENCE}
{Zhan \MakeLowercase{\textit{\etal}}: Multimodal Image Synthesis and Editing: A Survey and Taxonomy}

\IEEEtitleabstractindextext{
\begin{abstract}
As information exists in various modalities in real world, effective interaction and fusion among multimodal information plays a key role for the creation and perception of multimodal data in computer vision and deep learning research. With superb power in modeling the interaction among multimodal information, multimodal image synthesis and editing has become a hot research topic in recent years. Instead of providing explicit guidance for network training, multimodal guidance offers intuitive and flexible means for image synthesis and editing. On the other hand, this field is also facing several challenges in alignment of multimodal features, synthesis of high-resolution images, faithful evaluation metrics, etc. In this survey, we comprehensively contextualize the advance of the recent multimodal image synthesis and editing and formulate taxonomies according to data modalities and model types. We start with an introduction to different guidance modalities in image synthesis and editing, and then describe multimodal image synthesis and editing approaches extensively according to their model types. After that, we describe benchmark datasets and evaluation metrics as well as corresponding experimental results. Finally, we provide insights about the current research challenges and possible directions for future research. A project associated with this survey is available at \href{https://github.com/fnzhan/Generative-AI}{https://github.com/fnzhan/Generative-AI}.
\end{abstract}

}

\maketitle

\IEEEdisplaynontitleabstractindextext

\IEEEpeerreviewmaketitle

\IEEEraisesectionheading{\section{Introduction}\label{sec:introduction}}

\IEEEPARstart{H}{umans} are naturally capable of imaging a scene according to a piece of visual, text or audio description.
However, the intuitive processes are less straightforward for deep neural networks, primarily due to an inherent modality gap.
This \textit{modality gap} for visual perception can be boiled down to \textit{intra-modal gap} between visual clues and real images, and \textit{cross-modal gap} between non-visual clues and real images.
Targeting to mimic human imagination and creativity in the real world, the tasks of \textbf{M}ultimodal \textbf{I}mage \textbf{S}ynthesis and \textbf{E}diting (\textbf{MISE}) provide profound insights about how deep neural networks correlate multimodal information with image attributes.

\begin{figure*}[t]
\centering
\includegraphics[width=1.0\linewidth]{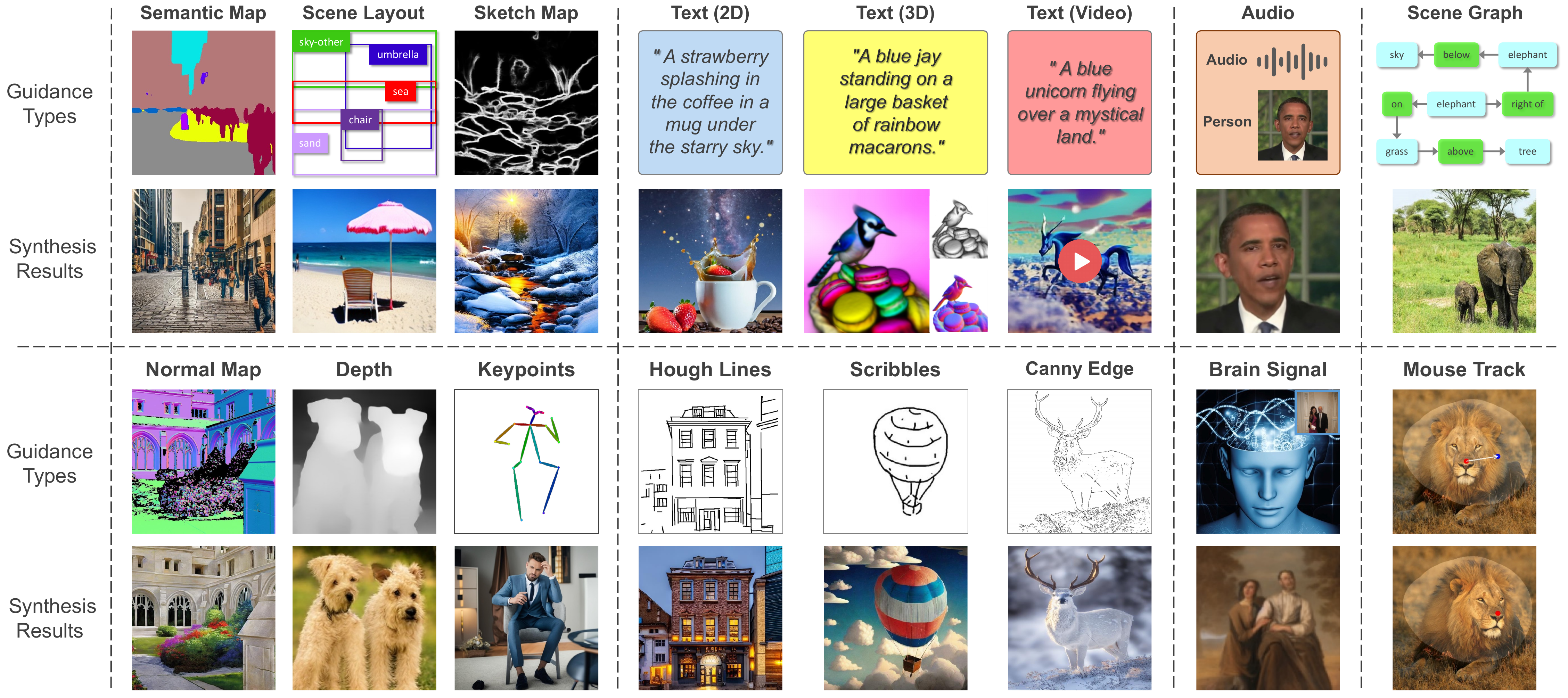}
\caption{
Illustration of multimodal image synthesis and editing. Typical guidance types include visual information (\eg  semantic maps, scene layouts, sketch maps), text prompts, audio signal, scene graph, brain signal, and mouse track.
The samples are from \cite{zhang2023adding,cheng2023layoutdiffuse,saharia2022photorealistic,poole2023dreamfusion,singer2022make,guo2021ad,li2019pastegan,esser2020taming,takagi2023improving,pan2023drag}.
}
\label{im_modality}
\end{figure*}

As a trending area, image synthesis and editing aim to create realistic images or edit real images with natural textures.
In the last few years, it has witnessed very impressive progress thanks to the advance of generative AI especially deep generative models \cite{goodfellow2014generative,esser2020taming,ho2020denoising} and neural rendering \cite{mildenhall2020nerf}.
To achieve controllable generation, a popular line of research focuses on generating and editing images conditioned on certain guidance as illustrated in Fig. \ref{im_modality}.
Typically, visual clues, such as segmentation maps and sketch maps, have been widely adopted to guide image synthesis and editing \cite{isola2017image,park2019semantic,lee2020maskgan}.
Beyond the intra-modal guidance of visual clues, cross-modal guidance such as texts, audios, and scene graph provides an alternative but often more intuitive and flexible way of expressing visual concepts.
However, effective retrieval and fusion of heterogeneous information from data of different modalities present a substantial challenge in multimodal image synthesis and editing.

As a pioneering effort in multimodal image synthesis, \cite{mansimov2015generating} shows that recurrent variational auto-encoder could generate novel visual scenes conditioned on image captions.
The research of multimodal image synthesis is then significantly advanced with the prosperity of Generative Adversarial Networks (GANs) \cite{goodfellow2014generative,mirza2014conditional,park2019semantic,isola2017image,shrivastava2017learning,chung2017you} and diffusion models \cite{dhariwal2021diffusion,ho2020denoising,song2020denoising,song2020score,jolicoeur2020adversarial}.
Originating from the Conditional GANs (CGANs) \cite{mirza2014conditional}, 
a bunch of GANs and diffusion models \cite{park2019semantic,isola2017image,ramesh2022hierarchical,saharia2022photorealistic} have been developed to synthesize images from various multimodal signals, by incorporating the multimodal guidance to condition the generation process.
This conditional generation paradigm is relatively straight-forward and is widely adopted in SOTA methods to yield unprecedented generation performance \cite{park2019semantic,zhou2021lafite,ho2022classifier,ramesh2022hierarchical,saharia2022photorealistic}.
On the other hand, developing conditional model require a cumbersome training process which usually involves high computational cost.
Thus, another line of research refers to pre-trained models for MISE, which can be achieved by manipulation in the GAN latent space via inversion \cite{xia2021tedigan,patashnik2021styleclip,xia2022gan,zhu2020domain}, applying guidance functions \cite{dhariwal2021diffusion,liu2021more} to diffusion process or adapting the latent space \& embedding \cite{kim2021diffusionclip,hertz2022prompt,ruiz2022dreambooth,gal2022image} of diffusion models.

Currently, a CNN architecture is still widely adopted in GANs and diffusion models, which hinders them from supporting diverse multimodal input in a unified manner.
On the other hand, with the prevalence of Transformer model \cite{vaswani2017attention} which naturally allows various multimodal input, impressive improvements have been made in the generation of different modality data, such as language \cite{radford2019language}, image \cite{chen2020generative}, and audio \cite{dhariwal2020jukebox}.
These recent advances fueled by Transformer suggest a possible route for autoregressive models \cite{gregor2014deep} in MISE by accommodating the long-range dependency of sequences.
Notably, both multimodal guidance and images can be represented in a common form of discrete tokens.
For instance, texts can be naturally denoted by token sequence; audio and visual guidance including images can be represented as token sequences \cite{oord2017neural}.
With such unified discrete representation, the correlation between multimodal guidance and images can be well accommodated via Transformer-based autoregressive models which have pushed the boundary of MISE significantly \cite{ramesh2021dalle,esser2020taming,wu2021nuwa}.

Most aforementioned methods work for 2D images regardless the 3D essence of real world.
With the recent advance of neural rendering, especially Neural Radiance Fields (NeRF) \cite{mildenhall2020nerf}, 3D-aware image synthesis and editing have attracted increasing attention from the community.
Distinct from synthesis and editing on 2D images, 3D-aware MISE poses a bigger challenge thanks to the lack of multi-view data and requirement of multi-view consistency during synthesis and editing.
As a remedy, pre-trained 2D foundation models (e.g., CLIP \cite{radford2021learning} and Stable Diffusion \cite{rombach2022high}) can be employed to drive the NeRF optimization for view synthesis and editing \cite{poole2023dreamfusion,lin2022magic3d}.
Besides, generative models like GAN and diffusion models can be combined with NeRF to train 3D-aware generative models on 2D images, where MISE can be performed by developing conditional NeRFs or inverting NeRFs \cite{sun2022ide,deng20233d}.

The contributions of this survey can be summarized in the following aspects:

\noindent
$\bullet$ This survey covers extensive literature with regard to multimodal image synthesis and editing with a rational and structured framework.

\noindent
$\bullet$ We provide a foundation of different types of guidance modality underlying multimodal image synthesis and editing tasks and elaborate the specifics of encoding approaches associated with the guidance modalities.

\noindent
$\bullet$ 
We develop a taxonomy of the recent approaches according to the essential models and highlight the major strengths and weaknesses of existing models.

\noindent
$\bullet$ This survey provides an overview of various datasets and evaluation metrics in multimodal image synthesis and editing, and critically evaluates the performance of contemporary methods.

\noindent
$\bullet$ We summarize the open challenges in the current research and share our humble opinions on promising areas and directions for future research.

The remainder of this survey is organized as follows. Section \ref{foundations} presents the modality foundations of MISE.
Section \ref{methods} provides a comprehensive overview and description of MISE methods with detailed pipelines.
Section \ref{experiments} reviews the common datasets and evaluation metrics, with experimental results of typical methods.
In Section \ref{future}, we discuss the main challenges and future research directions for MISE. 
Some social impact analysis and concluding remarks are drawn in Section \ref{social} and Section \ref{conclusion}, respectively.

\renewcommand\arraystretch{1.1}
\begin{table*}[t]
\caption
{
The strength and weakness of different model types for MISE tasks. Representative MISE works are also listed as references.
}
\renewcommand\tabcolsep{2.5pt}
\centering
\scriptsize
\begin{tabular}{llll}
\hline
\textbf{Model Types}  & \textbf{Strength}               & \textbf{Weakness}                 & \textbf{References}                 \\ \hline
  \multirow{4}{*}{\begin{tabular}[l]{@{}l@{}}  \textbf{GAN \cite{goodfellow2014generative}} \end{tabular}}&
  \multirow{4}{*}{\begin{tabular}[l]{@{}l@{}} Explicit latent space \\ Fast inference speed \end{tabular}} &
  \multirow{4}{*}{\begin{tabular}[l]{@{}l@{}} Instable training \\ Mode collapse \end{tabular}} &
                      \textbf{Visual:} Pix2PixHD\cite{wang2018high}, DIRT\cite{lee2018diverse}, SPADE\cite{park2019semantic}, OASIS\cite{sushko2020you}  \\  
                      &                                 &                                   & 
                      \begin{tabular}[l]{@{}l@{}} \textbf{Text:} StackGAN\cite{zhang2017stackgan}, AttnGAN\cite{xu2018attngan}, MirrorGAN\cite{qiao2019mirrorgan}, StyleCLIP\cite{patashnik2021styleclip}, GigaGAN\cite{kang2023scaling} \end{tabular}\\  
                      &                                 &                                   & 
                      \textbf{Audio:} Wav2Lip\cite{prajwal2020lip}, MakeitTalk\cite{zhou2020makelttalk}, PC-AVS\cite{zhou2021pose}, Synctalkface\cite{park2022synctalkface} \\  
                      &                                 &                                   & 
                      \textbf{Others:} PasteGAN\cite{li2019pastegan}, DragGAN\cite{pan2023drag}        \\ \hline
\multirow{4}{*}{\begin{tabular}[l]{@{}l@{}} \textbf{Diffusion} \\ \textbf{Model \cite{ho2020denoising}} \end{tabular}} &
  \multirow{4}{*}{\begin{tabular}[l]{@{}l@{}} Strong modeling capability\\ Stable training \\ Stationary training objective \end{tabular}} &
  \multirow{4}{*}{\begin{tabular}[l]{@{}l@{}} Unclear latent space \\ Intensive computation \\ Slow inference speed\end{tabular}} &
                      \textbf{Visual:} ControlNet\cite{zhang2023adding}, SDM \cite{wang2022semantic}, UniControl\cite{qin2023unicontrol} \\  
                      &                                 &                                   & 
                      \begin{tabular}[l]{@{}l@{}}\textbf{Text:} Imagen\cite{saharia2022photorealistic}, DALLE 2\cite{ramesh2022hierarchical}, LDM\cite{rombach2022high},  InstructPix2Pix\cite{brooks2023instructpix2pix}, DreamBooth\cite{ruiz2022dreambooth} \end{tabular}
                       \\ 
                       % DiffEdit\cite{couairon2022diffedit},
                      &                                 &                                   & \textbf{Audio:} DiffTalk\cite{shen2023difftalk}, EDGE \cite{tseng2023edge}, MM-Diffusion \cite{ruan2023mm} \\
                      &                                 &                                   & 
                      \textbf{Others:} SGDiff \cite{yang2022diffusion}, Takagi \& Nishimoto \cite{takagi2023improving} \\ \hline
\multirow{3}{*}{\begin{tabular}[l]{@{}l@{}} \textbf{Autoregressive} \\ \textbf{Model \cite{esser2020taming}} \end{tabular}} &
  \multirow{3}{*}{\begin{tabular}[l]{@{}l@{}}Natural multimodal support \\ Unified architecture \\ Stable training \end{tabular}} &
  \multirow{3}{*}{\begin{tabular}[l]{@{}l@{}} Lack of latent space \\ Slow inference speed\end{tabular}} &
                      \textbf{Visual:} ImageGPT \cite{chen2020generative}, Taming Transformer \cite{esser2020taming}, NÜWA \cite{wu2021nuwa}, Instaformer \cite{kim2022instaformer} \\  
                      &                                 &                                   & 
                      \begin{tabular}[l]{@{}l@{}}\textbf{Text:} CogView \cite{ding2021cogview}, DALL-E \cite{ramesh2021zero}, NÜWA \cite{wu2021nuwa}, Parti \cite{yu2022scaling}, Make-A-Scene \cite{gafni2022make}, Muse \cite{chang2023muse} \end{tabular}  \\ 
                      &                                 &                                   & 
                      \textbf{Audio:} Live speech portraits \cite{lu2021live}, AI Choreographer \cite{li2021ai}, Bailando \cite{siyao2022bailando} \\ \hline
                      % &                                 &                                   & 
                      % \textbf{Other:}                             \\ \hline
\multirow{3}{*}{\begin{tabular}[l]{@{}l@{}} \textbf{NeRF \cite{mildenhall2020nerf}} \end{tabular}}  & 
\multirow{3}{*}{\begin{tabular}[l]{@{}l@{}} 3D-consistency \end{tabular}}& 
\multirow{3}{*}{\begin{tabular}[l]{@{}l@{}} Limited multiview data \\ High computation cost \\ Inferior quality \end{tabular}} & 
\textbf{Visual:} OR-NeRF \cite{yin2023or}, SKED \cite{mikaeili2023sked}, SINE \cite{bao2023sine}, NeRF object removal \cite{weder2023removing}, SinNeRF \cite{xu2022sinnerf} \\  
                      &  &   & 
                      \textbf{Text:} LENeRF \cite{hyung2023local}, DreamFusion \cite{poole2023dreamfusion},  Magic3D \cite{lin2022magic3d}, CLIP-NeRF \cite{wang2022clip}, ProlificDreamer \cite{wang2023prolificdreamer} \\ 
                      &                                 &                                  & 
                      \textbf{Audio:} Geneface \cite{ye2023geneface}, Geneface++ \cite{ye2023geneface++}, AD-NeRF \cite{guo2021ad}, DFRF \cite{shen2022learning}, SSP-NeRF \cite{liu2022semantic}                       \\ \hline 
                      % &                                 &                                   & 
                      % Other:                             \\ \hline
\end{tabular}
\label{tab_models}
\end{table*}

% \vspace{15pt}

\section{Modality Foundations}
\label{foundations}

Each source or form of information can be called a modality. For example, people have the sense of touch, hearing, sight, and smell; the medium of information includes voice, video, text, etc.; data are recorded by various sensors such as radar, infrared, and accelerometer.
In terms of image synthesis and editing, we group the modality guidance as visual guidance, text guidance, audio guidance, and other modality guidance.
Detailed description of each modality guidance together with related processing methods will be presented in the following subsections.

\subsection{Visual Guidance}

Visual guidance has drawn widespread interest in the field of MISE due to its inherent capacity to convey spatial and structural details.
Notably, it encapsulates specific image properties in pixel space, thereby offering an exceptional degree of control. This property of visual guidance facilitates interactive manipulation and precise handling during image synthesis, which can be crucial for achieving desired outcomes. 
As a pixel-level guidance, it can be seamlessly integrated into the image generation process, underscoring its versatility and extensive use in various image synthesis contexts.
Common types of visual guidance encompass segmentation maps \cite{park2019semantic,isola2017image}, keypoints \cite{ma2017pose,men2020controllable,zhang2021deep}, 
sketch \& edge \& scribbles \cite{zhu2017toward,lee2018diverse,gao2020sketchycoco,chen2018sketchygan,chen2020deepfacedrawing,zhu2019deep,zhu2021learning,zhu2020knowledge,li2020staged}, and scene layouts \cite{sun2019lostgan,zhao2019layout2im,li2020bachgan,li2021image,frolov2021attrlostgan} as illustrated in Fig. \ref{im_modality}.
Besides, several studies investigate image synthesis conditioned on depth map \cite{esser2020taming,zhang2023adding}, normal map \cite{zhang2023adding}, trace map \cite{koh2021text}, etc.
The visual guidance can be obtained by employing pre-trained models (e.g., segmentation model, depth predictor, pose predictor), applying algorithms (e.g., Canny edges, Hough lines), or relying on manual effort (e.g., manual annotation, human scribbles).
By modifying the visual guidance elements, like semantic maps, we can directly repurpose image synthesis techniques for various image editing tasks \cite{zhan2022bi,zheng2022semantic}, demonstrating the versatile applicability of visual guidance in the domain of MISE.

\textbf{Visual Guidance Encoding.}
These visual cues, represented in 2D pixel space, can be interpreted as specific types of images, thereby permitting their direct encoding via numerous image encoding strategies such as Convolutional Neural Networks (CNNs) and Transformers.
As the encoded features spatially align with image features, it can be smoothly integrated into networks via naive concatenation, SPADE \cite{park2019semantic}, cross-attention mechanism \cite{rombach2022high}, etc.

\subsection{Text Guidance}

Compared with visual guidance, text guidance provides a more versatile and flexible way to express and describe visual concepts. This is because text can capture a wide range of ideas and details that may not be easily communicated through other means.
Text descriptions can be ambiguous and open to interpretation. This is both a challenge and an opportunity. It's a challenge because it can lead to a wide array of possible images that accurately represent the text, making it harder to predict the outcome. However, it's also an opportunity because it allows for greater creativity and diversity in the resulting images.
The text-to-image synthesis task \cite{zhang2017stackgan,zhang2018stackgan++,reed2016generative,frolov2021adversarial} aims to produce clear, photo-realistic images with high semantic relevance to the corresponding text guidance.
Notably, text and images are different types of data which makes it difficult to learn an accurate and reliable mapping from one to the other. Techniques for integrating text guidance, such as representation learning, play a crucial role in text-guided image synthesis and editing.

\textbf{Text Guidance Encoding.}
Learning faithful representation from text description is a non-trivial task.
There are a number of traditional text representations, such as Word2Vec \cite{mikolov2013distributed} and Bag-of-Words \cite{harris1954distributional}.
With the prevalence of deep neural networks, Recurrent Neural Network (RNN) \cite{reed2016generative} and LSTM \cite{xu2018attngan} are widely adopted to encode texts as features \cite{qiao2019mirrorgan}.
With the development of pre-trained models in natural language processing field,
several studies \cite{wang2021faces,pavllo2020controlling} also explore to perform text encoding by leveraging large-scale pre-trained language models such as BERT \cite{devlin2018bert}.
Remarkably, with a large number of image-text pairs for training, Contrastive Language-Image Pre-training (CLIP) \cite{radford2021learning} yields informative text embeddings by learning the alignment of images and the corresponding captions, and has been widely adopted for text encoding.

\subsection{Audio Guidance}
Unlike text and visual guidance, audio guidance provides temporal information which can be utilized for generating dynamic or sequential visual content.
The relationship between audio signals and images \cite{harwath2017learning,harwath2018vision,li2020direct} is often more abstract compared to text or visual guidance.
For instance, audio associated with certain actions or environments may suggest but not explicitly define visual content \cite{aytar2016soundnet}; sound can carry emotional tone and nuanced context that isn't always clear in text or visual inputs.
Thus, audio-guided MISE offers an interesting challenge of interpreting audio signals into visual content. This involves understanding and modeling the complex correlations between sound and visual elements, which has been explored in talking-face generation \cite{chen2019hierarchical,prajwal2020lip,zhou2021pose,park2022synctalkface} whose goal is to create realistic animations of a face speaking given an audio input.

\textbf{Audio Guidance Encoding.}
An audio sequence can be generated from given videos where deep convolution network is employed to extract features from video screenshots followed by LSTM \cite{hochreiter1997long} to generate audio waveform of the corresponding input video \cite{owens2016visually}.
Besides, an input audio segment can also be represented by a sequence of features which can be spectrograms, fBanks, Mel-Frequency Cepstral Coefficients (MFCCs), and the hidden layer outputs of the pre-trained SoundNet model \cite{aytar2016soundnet}.
In talking face generation \cite{song2018talking}, Action Units (AUs) \cite{ekman2002facial} has also been widely adopted to convert the driving audio into coherent visual signals for talking face generation.

\subsection{Other Modality Guidance}

Several other types of guidance have also been investigated to guide multimodal image synthesis and editing.

\textbf{Scene Graph.}
Scene Graphs represent scenes as directed graphs, where nodes are objects and edges give relationships between objects. Image generation conditioned on scene graphs allows to reason explicit object relationships and synthesize faithful images with complex scene relationships. The guided scene graph can be encoded through a graph convolution network \cite{johnson2018image} which predicts object bounding boxes to yield a scene layout. 
For instance, 
Vo \etal \cite{vo2020visual} propose to predict relation units between objects which is converted to a visual layout via convolutional LSTM \cite{shi2015convolutional}.

\textbf{Brain Signal.}
Treating brain signals as a modality to synthesize or reconstruct visual images offers an exciting way to understand brain activity and facilitate brain-computer interfaces. Recently, several studies explore to generate images from functional magnetic resonance imaging (fMRI). For example, Fang \etal \cite{fang2020reconstructing} decode shape and semantic representations from the visual cortex, and then fuse them to generate images via GAN; Lin \etal \cite{lin2022mind} propose to map fMRI signals into the latent space of pretrained StyleGAN to enable conditional generation; Takagi and Nishimoto \cite{takagi2022high} quantitatively interpret each component in pretrained LDM \cite{rombach2022high} by mapping them into distinct brain regions.

\textbf{Mouse Track.}
To achieve precise and flexible manipulation of image content, mouse track \cite{pan2023drag} has recently emerged as a remarkable guidance in MISE. Specifically, users can select a set of `handle points' and `target points' within an image by simply clicking the mouse. The objective here is to edit the image by steering these handle points to their respective target points.
This innovative approach of mouse track guidance enables an image to be deformed with an impressive level of accuracy, and facilitates manipulation of various attributes such as pose, shape, and expression across a range of categories.
The point motion can be integrated to supervise the editing via a pre-trained transformer that's based on optical flow \cite{yang2020upgrading,endo2022user} or a shifted patch loss on the generator features \cite{pan2023drag}.

% \vspace{15pt}

\section{Methods}
\label{methods}

We broadly categorize the methods for MISE into five categories: GAN-based methods (Sec.~\ref{gan}),
autoregressive methods (Sec.~\ref{autoregressive}), 
diffusion-based methods (Sec.~\ref{diffusion}),
NeRF-based methods (Sec.~\ref{nerf}),
and other methods (Sec.~\ref{others}).
We briefly summarize the strength and weakness of four main methods with representative references as shown in Table~\ref{tab_models}.
In this section, we first discuss the GAN-based methods, which generally rely on GANs and their inversion. We then discuss the prevailing diffusion-based methods and autoregressive methods comprehensively.
After that, we introduce NeRF for the challenging task of 3D-aware MISE.
Later, we present several other methods for image synthesis and editing under the context of multimodal guidance.
Finally, we compare and discuss the strengths and weaknesses of different generation architectures.

\begin{figure}[t]
\centering
\includegraphics[width=1.0\linewidth]{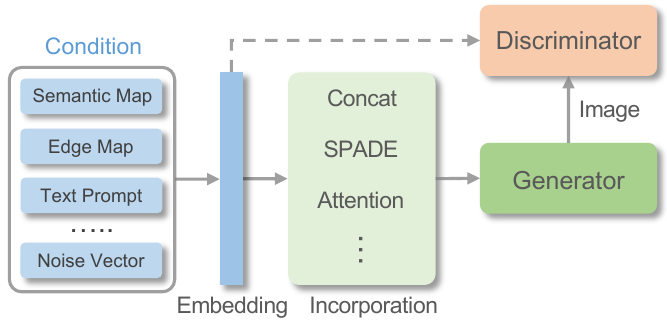}
\caption{
Illustration of conditional GAN framework with different condition incorporation mechanisms.
}
\label{gan_incorporation}
\end{figure}

\subsection{GAN-based Methods}
\label{gan}

GAN-based methods have been widely adopted for various MISE tasks by either developing conditional GANs (Sec. \ref{conditional_gans}) or leveraging pre-trained unconditional GANs (Sec. \ref{inversion}).
For conditional GANs, multimodal condition can be directly incorporated into the generator to guide the generation process.
For pre-trained unconditional GANs, GAN inversion is usually employed to perform various MISE tasks by operating latent codes in latent spaces.

\subsubsection{Conditional GANs}
\label{conditional_gans}
Conditional Generative Adversarial Networks (CGANs) \cite{mirza2014conditional} are extensions of the popular GAN architecture which allow for image generation with specific characteristics or attributes.
The key idea behind CGANs is to condition the generation process on additional information, such as multimodal guidance in MISE tasks. This is achieved by feeding the additional information into both the generator and discriminator networks as extra guidance. The generator then learns to generate samples that not only fool the discriminator but also match the specified conditional information.
In recent years, a range of designs have significantly boosted the performance of CGANs for MISE \cite{frolov2021adversarial} \footnote{Please refer to \cite{frolov2021adversarial} for detailed review of GAN-based text-to-image generation.}.

\textbf{Condition Incorporation.}
To steer the generation process, it is necessary to incorporate multimodal conditions into the network effectively as shown in Fig. \ref{gan_incorporation}.
Generally, multimodal guidance can be uniformly encoded as 1-D features which can be concatenated with the feature in networks \cite{mirza2014conditional,reed2016generative,zhou2021pose}.
For visual guidance that is spatially aligned with the target image, the condition can be directly encoded as 2D features which provide accurate spatial guidance for generation or editing \cite{isola2017image}.
However, the encoded 2D features struggle to capture complex scene structural relationships between the guidance and real images when there exists very different views or severe deformations.
Under such circumstances, an attention module can be employed to align the guidance with the target image as in \cite{tang2019multi,zhang2020cross,zhan2021unite}.
Moreover, naively encoding the visual guidance with deep networks is suboptimal as part of the guidance information tends to be lost in normalization layers.
Thus, a spatially-adaptive de-normalization (SPADE) \cite{park2019semantic} is introduced to inject the guided feature effectively, which is further extended to a semantic region-adaptive normalization \cite{zhu2020sean} to achieve region-wise condition incorporation.
Besides, by assessing the similarity between generated images and conditions, an attentional incorporation mechanism \cite{xu2018attngan,tan2019semantics,li2019controllable,zhu2019dm} can be employed to direct the generator's attention to particular image regions during generation,
which is particularly advantageous when dealing with complex conditional information, such as texts.
Notably, complex conditions also can be mapped to an intermediary representation which facilitates more faithful image generation, \eg, audio clip can be mapped to facial landmarks \cite{chen2019hierarchical,zhou2020makelttalk} or 3DMM parameters \cite{blanz1999morphable} for talking-face generation.
For sequential conditions such as audios \cite{chung2017you,song2018talking,chen2019hierarchical,chen2020talking,zhou2019talking,zhou2021pose,suwajanakorn2017synthesizing,wang2021one}, a recurrent condition incorporation mechanism is also widely adopted to account for temporal dependency such that smooth transition can be achieved in sequential conditions.

\begin{figure*}[t]
\centering
\includegraphics[width=1.0\linewidth]{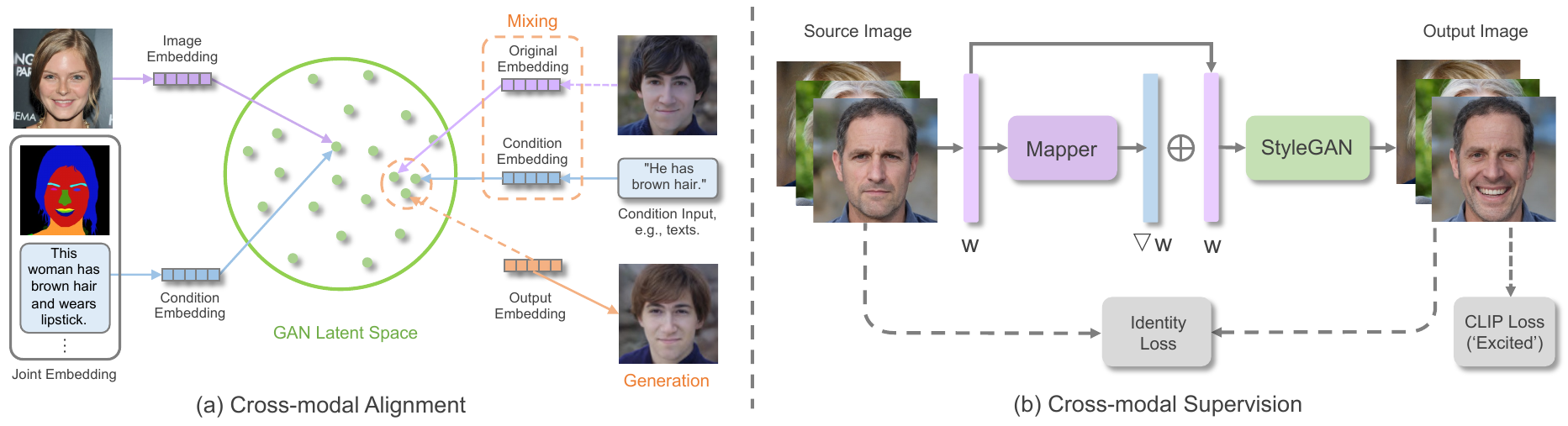}
\caption{
The architectures of GAN inversion for MISE, including (a) Cross-modal alignment \cite{xia2021tedigan} and (b) cross-modal supervision \cite{patashnik2021styleclip}.
The cross-modal alignment embeds both images and conditions into the latent space of GAN (\eg, StyleGAN \cite{karras2019style}), aiming to pull their embeddings to be closer.
Then image and condition embeddings can be mixed to perform multimodal image generation or editing.
The cross-modal supervision inverts source image into a latent code and trains a mapper network to produce residuals that are added to the latent code to yield the target code, from which a pre-trained StyleGAN generates an image assessed by the CLIP and identity losses.
The figure is reproduced based on \cite{xia2021tedigan} and \cite{patashnik2021styleclip}.
}
\label{gan_inversion}
\end{figure*}

\textbf{Model Structure.}
Conditional generation of high-resolution images with fine details is challenging and computationally expensive for GANs. 
Coarse-to-fine structures \cite{zhang2017stackgan,zhang2018stackgan++,wang2018high,karras2017progressive,li2020staged} help address these issues by gradually refining the generated images or features from low resolutions to high resolutions. 
By generating coarse images or features first and then refining them, the generator network can focus on capturing the overall structure of the image before moving on to the fine details, which leads to more efficient training and higher generation quality.
Not only generator, many discriminator networks \cite{zhang2018photographic,wang2018high} also operate at multiple levels of resolution to efficiently differentiate high-resolution images and avoid potentially overfitting.
On the other hand, as a scene can be depicted with diverse linguistic expressions, generating images with consistent semantics regardless of the expression variants presents a significant challenge.
Multiple pieces of research employ a siamese structure with two generation branches to facilitate the semantic alignment. With a pair of conditions for the two branches, a contrastive loss can be adopted to minimize the distance between positive pairs (two text prompts describe the same scene) and maximize the distance between negative pairs (two prompts describe different scenes) \cite{tan2019semantics,yin2019semantics,cha2019adversarial}.
Besides, an intra-domain transformation loss \cite{amodio2019travelgan} can also be employed in siamese structure to preserve key characteristics during generation.
Except for above structures, a cycle structure also has been explored in series of conditional GANs to preserve key information in generation process.
Specifically, some research \cite{zhu2017unpaired,tang2019cycle,lao2019dual,chen2019cycle,qiao2019mirrorgan} explores to pass the generated images through an inverse network to yield the conditional input, which imposes a cycle-consistency of conditional input.
The inverse network varies for different conditional inputs, \eg, image captioning models \cite{nguyen2017plug,qiao2019mirrorgan} for text guidance, generation networks for visual guidance.

\textbf{Loss Design.}
Except for the inherent adversarial loss in GANs, various other loss terms have been explored to achieve high-fidelity generation or faithful conditional generation.
For conditional input that is spatially aligned with the ground-truth image, it has been proved that perceptual loss \cite{johnson2016perceptual} is able to boost the generation quality significantly \cite{wang2018perceptual}, by minimizing the distance of perceptual features between generated images and the ground-truth.
Besides, associated with the cycle structure described previously, a cycle-consistency loss \cite{zhu2017unpaired} is duly imposed to enforce condition consistency.
However, cycle-consistency loss is too restrictive for conditional generation as it assumes a bi-jectional relationship between two domains. Thus, some efforts \cite{benaim2017one,amodio2019travelgan,fu2019geometry} have been devoted to exploring one-way translation and bypass the bijection constraint of cycle-consistency.
With the emergence of contrastive learning, several studies explore to maximize the mutual information of positive pairs via noise contrastive estimation \cite{oord2018representation} for the preservation of contents in unpaired image generation from visual guidance \cite{park2020contrastive,andonian2021contrastive} or text-to-image generation \cite{zhang2021cross}.
Except for contrastive loss, triplet loss also has been employed to improve the condition consistency for cross-modal guidance like texts \cite{yin2019semantics}.

\subsubsection{Inversion of Unconditional GAN}
\label{inversion}
Large scale GANs~\cite{brock2018large,karras2019style} have achieved remarkable progress in unconditional image synthesis with high-resolution and high-fidelity.
With a pre-trained GAN model, a series of studies explore to invert a given image back into the latent space of the GAN, which is termed as GAN inversion \cite{xia2022gan} \footnote{Please refer to \cite{xia2022gan} for a comprehensive review of GAN inversion.}.
Specifically,
a pre-trained GAN learns a mapping from latent codes to real images, while the GAN inversion maps images back to latent codes, which is achieved by feeding the latent code into the pre-trained GAN to reconstruct the image through optimization.
Typically, the reconstruction metrics are based on $\ell_1$, $\ell_2$, perceptual~\cite{johnson2016perceptual} loss or LPIPS~\cite{zhang2018unreasonable}. Certain constraints on face identity~\cite{richardson2020encoding} or latent codes~\cite{zhu2020domain} could also be included during optimization.
With the obtained latent codes, we can faithfully reconstruct the original image and conduct realistic image manipulation in the latent space.
In terms of MISE, cross-modal image manipulation can be achieved by manipulating or generating latent codes according to the guidance from other modalities.

\textbf{Explicit Cross-modal Alignment.}
One direction of leveraging the guidance from other modalities is to map the embeddings of images and cross-modal inputs (\eg, semantic maps, texts) in a common embedding space \cite{xia2021tedigan,wang2021cycle} as shown in Fig. \ref{gan_inversion} (a). For example, TediGAN \cite{xia2021tedigan} trains an encoder for each modality to extract the embeddings and apply similarity loss to map them into the latent space. Afterwards, latent manipulation (\eg, latent mixing \cite{xia2021tedigan}) could be performed to edit the image latent codes toward the embeddings of other modalities and achieve cross-modal image manipulation. However, mapping multimodal data into a common space is non-trivial thanks to the heterogeneity across different modalities, which can result in inferior and unfaithful image generation.

\textbf{Implicit Cross-modal Supervision.}
Instead of explicitly projecting guidance modality into the latent space, another line of research aims to guide the synthesis or editing by defining consistency loss between the generation results and the guiding modality. For instance, Jiang \etal \cite{jiang2021talk} propose to optimize image latent codes through a pre-trained fine-grained attribute predictor, which can examine the consistency of the edited image and the text description. However, the attribute predictor is specifically designed for face editing with fine-grained attribute annotations, making it hard to generalize to other scenarios. 
A recently released large-scale pretrained model, Contrastive Language-Image Pre-training (CLIP) \cite{radford2021learning} has demonstrated great potential in multimodal synthesis and manipulation \cite{ramesh2021dalle, patashnik2021styleclip}, which learns joint vision-language representations from over 400M text-image pairs via contrastive learning. 
On the strength of the powerful pre-trained CLIP,
Bau \etal \cite{bau2021paint} define a CLIP-based semantic consistency loss to optimize latent codes inside an inpainting region to align the recovered content with the given text. Similarly, StyleClip \cite{patashnik2021styleclip} and StyleMC \cite{kocasari2021stylemc} employ cosine similarity between CLIP representations to supervise the text-guided manipulation as illustrated in Fig. \ref{gan_inversion} (b). A known issue of standard CLIP loss is the adversarial solution \cite{liu2021fusedream}, where the model tends to fool the CLIP classifier by adding meaningless pixel-level perturbations to the image. To this end, Liu \etal propose AugCLIP score \cite{liu2021fusedream} to robustify the standard CLIP score; StyleGAN-NADA \cite{gal2021stylegan} presents a directional CLIP loss to align the CLIP-space directions between the source and target text-image pairs. It also directly finetunes the pretrained generative model with text conditions for domain adaptation. Moreover, Yu \etal \cite{yu2022towards} introduce a CLIP-based contrastive loss for robust optimization and counterfactual image manipulation.

\subsection{Diffusion-based Methods}
\label{diffusion}

Recently, diffusion models such as denoising diffusion probabilistic models (DDPMs) \cite{ho2020denoising,sohl2015deep} have achieved great successes in generative image modeling \cite{ho2020denoising,song2020denoising,song2020score,jolicoeur2020adversarial}.
DDPMs are a type of latent variable models that consist of a forward diffusion process and a reverse diffusion process.
The forward process is a Markov chain where noise is gradually added to the data when sequentially sampling the latent variables $\x_t$ for $t = 1, \cdots, T$. Each step in the forward process is a Gaussian transition $q(\x_t | \x_{t-1}) :=\mathcal{N}(\sqrt{1-\beta_t}\x_{t-1}, \beta_t\mathbf{I})$, where $\{\beta_t\}^T_{t=0}$ are fixed or learned variance schedule. 
The reverse process $q(\x_{t-1} | \x_{t})$ is parameterized by another Gaussian transition
$p(\x_{t-1}|\x_{t}) := \mathcal{N}(\x_{t-1}; \mu(\x_t), \sigma_t^2\mathbf{I})$. 
${\mu(\x_t)}$ can be decomposed into a linear combination of $\x_t$ and a noise approximation model $\epsilon_{\theta}(\x_t, t)$ that can be learned through optimization. After training $\epsilon(\x, t)$, the sampling process of DDPM can be achieved by following a reverse diffusion process.

Song \etal \cite{song2020denoising} propose an alternative non-Markovian noising process that has the same forward marginals as DDPM but allows using different samplers by changing the variance of the noise. Especially, by setting the noise to 0, which is a DDIM sampling process \cite{song2020denoising}, the sampling process becomes deterministic, enabling full inversion of the latent variables into the original images with significantly fewer steps \cite{song2020denoising, dhariwal2021diffusion}.
Notably, the latest work \cite{dhariwal2021diffusion} has demonstrated even higher quality of image synthesis performance compared to variational autoencoders (VAEs) \cite{kingma2013auto}, flow models \cite{rezende2015variational,dinh2016density}, autoregressive models \cite{menick2018generating,van2016pixel} and (GANs) \cite{goodfellow2014generative,karras2019style}.
To achieve image generation and editing conditioned on provided guidance, leveraging pre-trained models \cite{liu2021more} (by guidance function or fine-tuning) and training conditional models from scratch \cite{rombach2022high} are both extensively studied in the literature.
A downside of guidance function method lies in the requirement of an additional guidance model which leads to a complicated training pipeline. 
Recently, Ho \etal \cite{ho2022classifier} achieve compelling results without a separately guidance model by using a form of guidance that interpolates between predictions from a diffusion model with and without labels.
GLIDE \cite{nichol2021glide} compares CLIP-guided diffusion model and conditional diffusion model on text-to-image synthesis task, and concludes that training conditional diffusion model yields better generation performance.

\begin{figure}[t]
\centering
\includegraphics[width=1.0\linewidth]{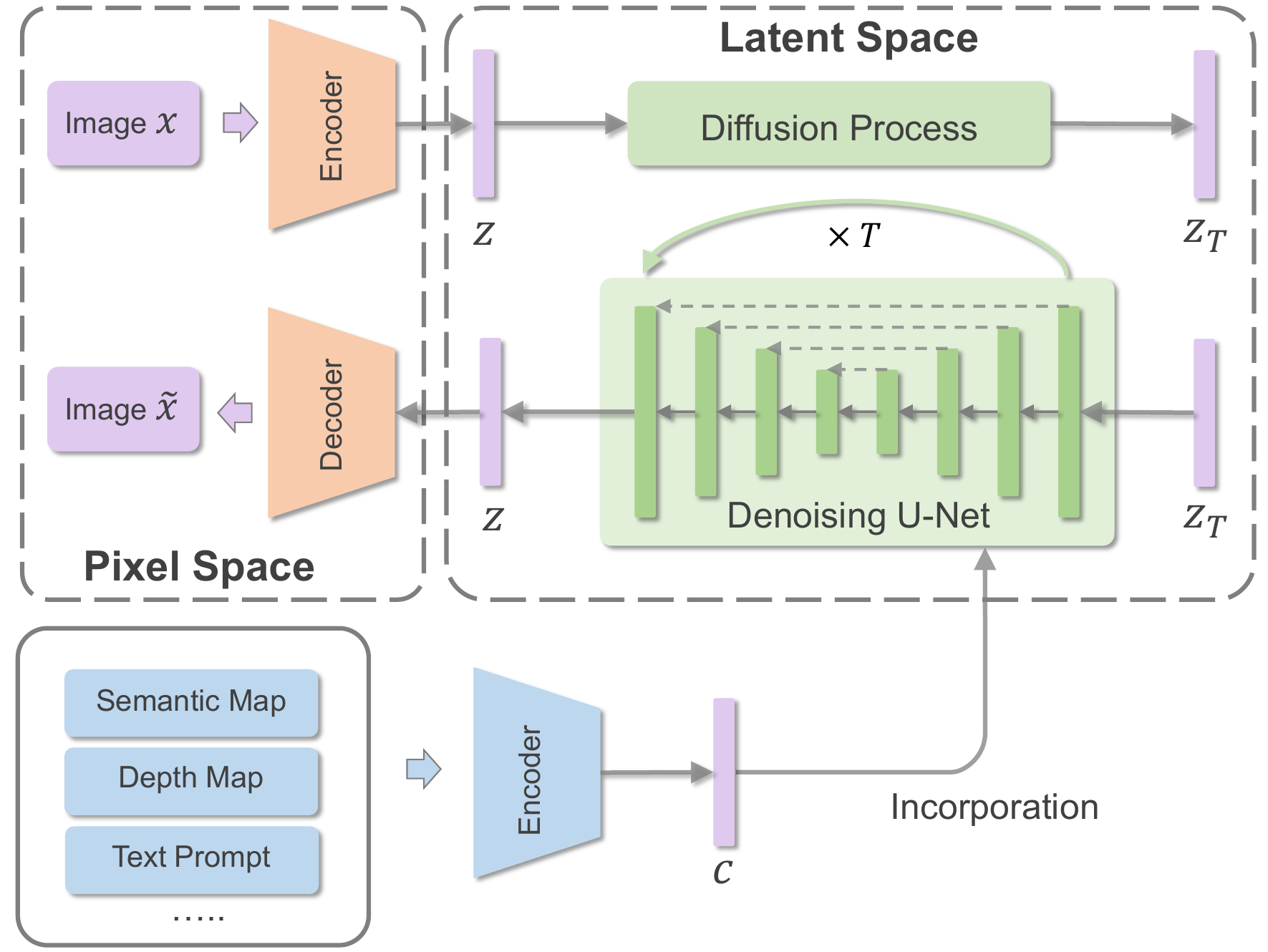}
\caption{
Overall framework of conditional diffusion model. 
With a certain model for latent representation, diffusion process models the latent space by reversing a forward diffusion process conditioned on certain guidance (e.g., semantic map, depth map, and texts).
The image is reproduced based on \cite{rombach2022high}.
}
\label{diffusion_conditional}
\end{figure}

\begin{figure*}[t]
\centering
\includegraphics[width=1.0\linewidth]{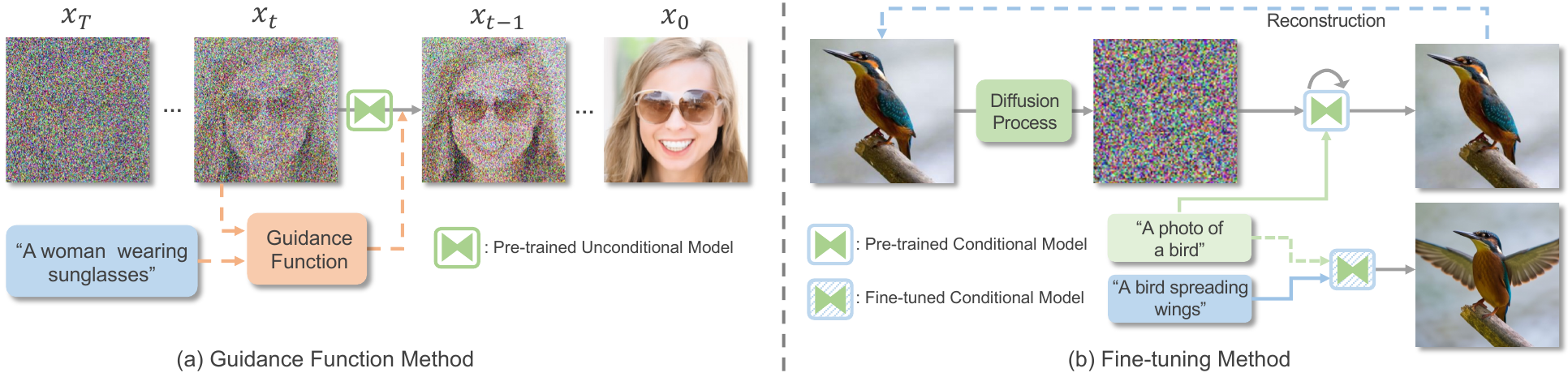}
\caption{
Typical frameworks of pre-trained diffusion models for MISE tasks, including guidance function method and fine-tuning method.
The figure is reproduced based on \cite{liu2021more} and \cite{kawar2022imagic}.
}
\label{diffusion_pretrained}
\end{figure*}

\subsubsection{Conditional Diffusion Models}
To launch the MISE tasks, a conditional diffusion model can be formulated by directly integrating the condition information into the denoising process.
Recently, the performance of conditional diffusion models is significantly pushed forward by a series of designs.

\textbf{Condition Incorporation.}
As a common framework, a condition-specific encoder is usually employed to project multimodal condition into embedding vectors, which is further incorporated into the model as shown in Fig. \ref{diffusion_conditional}. The condition-specific encoder can be learned along with the model or directly borrowed from pre-trained models.
Typically, CLIP is a common choice for text embedding as adopted in DALL-E 2 \cite{ramesh2022hierarchical}. Besides, generic large language models (\eg T5 \cite{raffel2020exploring}) pre-trained text corpora also show remarkable effectiveness at encoding text for image synthesis as validated in Imagen \cite{saharia2022photorealistic}.
With the condition embedding, diverse mechanisms can be adopted to incorporate it into diffusion models.
Specifically, the condition embedding can be naively concatenated or added to the diffusion timestep embedding \cite{dhariwal2021diffusion,ho2022cascaded}.
In LDM \cite{rombach2022high}, condition embedding is mapped to the intermediate layers of diffusion models via a cross-attention mechanism.
Imagen \cite{saharia2022photorealistic} further compares mean pooling and attention pooling with cross attention mechanism and observes both pooling mechanisms perform significantly worse.
To fully leverage the conditional information for semantic image synthesis, Wang \etal \cite{wang2022semantic} propose to incorporate visual guidance via a spatially-adaptive normalization, which improves both the quality and semantic coherence of generated images.
Instead of incorporating condition to train diffusion models from scratch, ControlNet \cite{zhang2023adding} aims to incorporate condition into a pre-trained diffusion model for controllable generation.
To preserve production-ready weights of pre-trained models for fast convergence, a `zero convolution' is designed to incorporate the guidance, where the convolution weights are gradually learned from zeros to optimized parameters.

\textbf{Latent Diffusion.}
To enable diffusion models training on limited computational resources while retaining their quality and flexibility,
several works explore to conduct diffusion process in learned latent spaces \cite{rombach2022high} as shown in Fig. \ref{diffusion_conditional}.
Typically, an autoencoding model can be employed to learn a latent space that is perceptually equivalent to the image space.
On the other hand, the learned latent spaces may be accompanied with undesired high variance, which highlighting the need for latent space regularizations. 
As a common choice, KL divergence can be applied to regularize the latent space towards a standard normal distribution.
Alternatively, vector quantization can also be applied for regularization via a VQGAN \cite{esser2020taming} variant with an absorbed quantization layer as in \cite{rombach2022high}.
Besides, VQGAN can directly learn a discrete latent space (quantization layer is not absorbed), which can be modeled by a discrete diffusion process as in VQ-Diffusion \cite{gu2021vector}.
Tang \etal \cite{tang2022improved} further improve VQ-Diffusion by introducing a high-quality inference strategy to alleviate the joint distribution issue.

\textbf{Model Architecture.}
Ho \etal \cite{ho2020denoising} introduced a U-Net architecture for diffusion models, which can incorporate the inductive bias of CNNs into the diffusion process.
This U-Net architecture is further improved by a series of designs, including attention configuration \cite{dhariwal2021diffusion}, residual block for upsampling and downsampling activations \cite{song2020score}, and adaptive group normalization \cite{dhariwal2021diffusion}.
Although U-Net structure is widely adopted in SOTA diffusion models, Chahal~\cite{chahal2022exploring} shows that a Transformer-based LDM \cite{rombach2022high} can yield comparable performance to U-Net-based LDM \cite{rombach2022high}, accompanied with a natural multimodal condition incorporation via multi-head attention.
Nevertheless, such Transformer architecture is more favored under the setting of discrete latent space as in \cite{jiang2022text2human,gu2021vector}.
On the other hand, instead of directly generating final images,
DALL-E 2 \cite{ramesh2022hierarchical} proposes a two-stage structure by producing intermediate image embeddings from text in the CLIP latent space.
Then, the image embeddings are applied to condition a diffusion model to generate final images, which allows improving the diversity of generated images \cite{ramesh2022hierarchical}.
Besides, some other architectures are also explored, including compositional architecture \cite{liu2022compositional} which generates an image by composing a set of diffusion models, multi-diffusion architecture \cite{bar2023multidiffusion} which is composed of multiple diffusion processes with shared parameters or constraints, retrieval-based diffusion model \cite{blattmann2022retrieval} which alleviates the high computational cost, etc.

\subsubsection{Pre-trained Diffusion Models}

Rather than expensively re-training diffusion models, another line of research resorts to guiding the denoising process with proper supervision, or finetuning the model with a lower cost as shown in Fig. \ref{diffusion_pretrained}.

\textbf{Guidance Function Method.}
As an early exploration, Dhariwal \etal \cite{dhariwal2021diffusion} augment pre-trained diffusion models with classifier guidance which can be extended to achieve conditional generation with various guidance.
Specifically, the reverse process $p(\x_{t-1}|\x_{t})$ with guidance can be rewritten as $p(\x_{t-1}|\x_{t},y)$ where $y$ is the provided guidance.
Following the derivation in \cite{dhariwal2021diffusion}, the final diffusion sampling process can be rewritten as:
\begin{equation}
    \x_{t-1} = \mu(\x_t) + \sigma_t^2 \nabla_{x_t} \log p(y|x_t) + \sigma_t \varepsilon, \ \ \varepsilon \sim \mathcal{N}(0, \mathbf{I})
\end{equation}
$F(x_t, y)=\log p(y|x_t)$ (dubbed as guidance function) indicates the consistency between $x_t$ and guidance $y$ which can be formulated by certain similarity metric \cite{liu2021more} such as Cosine similarity and L2 distance.
As the similarity is usually computed on the feature space, pre-trained CLIP can be adopted as the image encoder and condition encoder for text guidance as shown in Fig. \ref{diffusion_pretrained} (a).
However, the image encoder will take noisy images as input while CLIP is trained on clean images.
Thus, a self-supervised fine-tuning of CLIP can be performed to force an alignment between features extracted from clean and noised images as in \cite{liu2021more}.

To control the generation consistency with the guidance, a parameter $\gamma$ can be introduced to scale the guidance gradients as below:
\begin{equation}
    \x_{t-1} = \mu(\x_t) + \sigma_t^2 \gamma \nabla_{x_t} \log p(y|x_t) + \sigma_t \varepsilon, \ \ \varepsilon \sim \mathcal{N}(0, \mathbf{I})
\end{equation}
Apparently, the model will focus more on the modes of guidance with a larger gradient scale $\gamma$.
As the result, $\gamma$ is positively correlated with the generation consistency (with the guidance), while is negatively correlated with the generation diversity \cite{dhariwal2021diffusion}.
Besides, to achieve the local guidance for image editing,
a blended diffusion mechanism \cite{avrahami2021blended} can be employed by spatially blending noised image with the local guided diffusion latent at progressive noise levels.

\textbf{Fine-tuning Method.}
In terms of fine-tuning, MISE can be achieved by modifying the latent code or adapting the pre-trained diffusion models as shown in Fig. \ref{diffusion_pretrained} (b).
To adapt unconditional pre-trained models for text-guided editing, the input image is first converted to the latent space via the forward diffusion process. The diffusion model at the reverse path is then fine-tuned to generate images driven by the target text and the CLIP loss \cite{kim2021diffusionclip}.
For pre-trained conditional models (typically conditioned on texts), similar to GAN Inversion, a text latent embedding or a diffusion model can be fine-tuned to reconstruct a few images (or objects) \cite{ruiz2022dreambooth,gal2022image} faithfully. Then the obtained text embedding or fine-tuned model can be applied to generate the same object in novel contexts.
However, these methods \cite{ruiz2022dreambooth,gal2022image} usually drastically change the layout of the original images.
As observing the crux of the relationship between image spatial layout and each word lies in cross-attention layers, Prompt-to-Prompt \cite{hertz2022prompt} proposes to preserve some content from the original image by manipulating the cross-attention maps.
Alternatively, taking advantage of the step-by-step diffusion sampling process, a model fine-tuned for image reconstruction can be utilized to provide score guidance for content and structure preservation at the early stage of the denoising process \cite{zhang2022sine}.
Similar approach is adopted in \cite{kawar2022imagic} by fine-tuning diffusion model and optimizing text embedding via image reconstruction, which allows preserving contents via text embedding interpolation.

\begin{figure*}[t]
\centering
\includegraphics[width=0.99\linewidth]{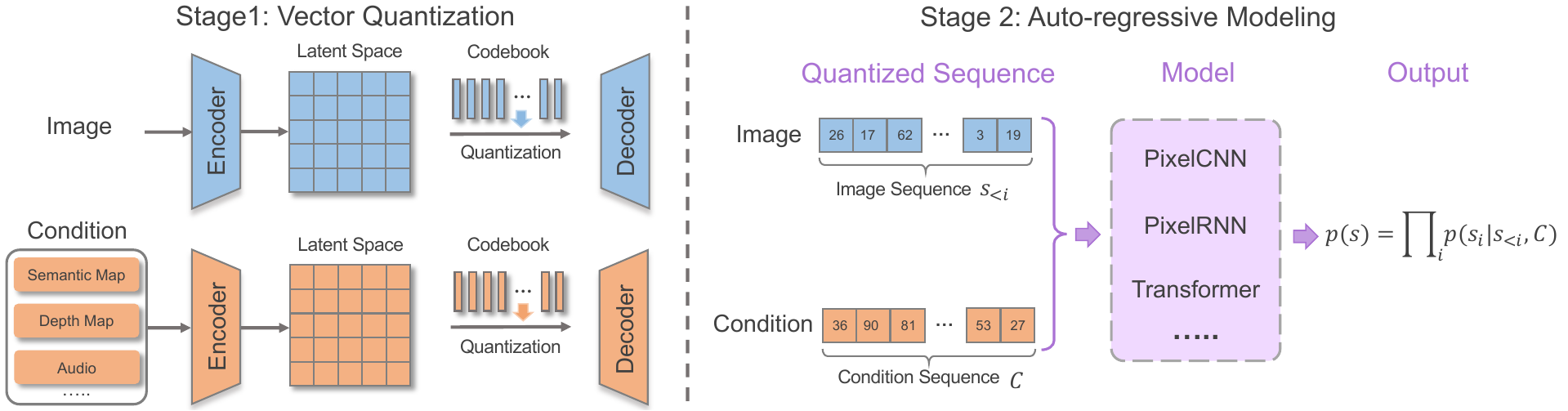}
\caption{
Typical framework of autoregressive methods for MISE tasks.
A quantization stage is firstly performed to learn discrete and compressed representation by reconstructing the original image or condition (e.g., semantic map) faithfully via VQ-GAN \cite{oord2017neural,esser2020taming},
followed by an autoregressive modeling stage to capture the dependency of discrete sequence.
The image is reproduced based on \cite{esser2020taming} and \cite{zhan2022auto}.
}
\label{ar_framework}
\end{figure*}

\subsection{Autoregressive Methods}
\label{autoregressive}

Fueled by the advance of GPT \cite{radford2019language} in natural language modeling, 
autoregressive models have been successfully applied to image generation \cite{chen2020generative} by treating the flattened image sequences as discrete tokens.
The plausibility of generated images demonstrates that autoregressive models are able to accommodate the spatial relationships between pixels and high-level attributes.
Compared with CNN, Transformer models naturally support various multimodal inputs in a unified manner, and a series of studies have been proposed to explore multimodal image synthesis with Transformer-based autoregressive models \cite{esser2020taming,ding2021cogview,esser2021imagebart,wu2021nuwa}.
Overall, the pipeline of autoregressive model for MISE consists of a \textit{vector quantization} \cite{oord2017neural,rolfe2016discrete} stage to yield unified discrete representation and achieve data compression, and an \textit{autoregressive modeling} stage which establishes the dependency between discrete tokens in a raster-scan order as illustrated in Fig. \ref{ar_framework}.

\subsubsection{Vector Quantization}

Directly treating all image pixels as a sequence for autoregressive modeling with Transformer is expensive in terms of memory consumption as the self-attention mechanism in Transformer incurs quadratic memory cost.
Thus, compressed and discrete representation of image is essential and significant for autoregressive image synthesis and editing.
A k-means method to cluster RGB pixel values has been adopted in \cite{chen2020generative} to reduce the input dimensionality.
However, k-means clustering only reduces the dimensionality while the sequence length is still unchanged. Thus, the autoregressive model still cannot be scaled to higher resolutions, due to the quadratically increasing cost in sequence length. To this end, Vector Quantised VAE (VQ-VAE) \cite{oord2017neural} is
adopted to learn discrete and compressed image representation.
VQ-VAE consists of an encoder, a feature quantizer, and a decoder.
The image is fed into the encoder to learn a continuous representation, which is quantized via the feature quantizer by assigning the feature to the nearest codebook entry.
Then the decoder reconstructs the original image from the quantized feature, driving to learn a faithful discrete image representation.
As assigning codebook entry is not differentiable, 
a reparameterization trick \cite{bengio2013estimating,oord2017neural} is usually adopted to approximate the gradient.
Targeting for learning superior discrete image representation, a series of efforts \cite{esser2020taming,yu2021vector,shin2021translation} have been devoted to improving VQ-VAE in terms of loss function, model architecture, codebook utilization, and learning regularization.

\textbf{Loss Function.}
To achieve desirable perceptual quality for reconstructed images, an adversarial loss and a perceptual loss \cite{johnson2016perceptual,lamb2016discriminative,larsen2016autoencoding} (with pre-trained VGG) can be incorporated for image reconstruction. 
With the extra adversarial loss and perceptual loss, the image quality is clearly improved compared with the original pixel loss as validated in \cite{esser2020taming}.
Except for pre-trained VGG for computing perceptual loss, vision Transformer \cite{dong2021peco} from self-supervised learning \cite{bao2021beit,devlin2018bert} is also proved to work well for calculating perceptual loss.
Besides, to emphasize reconstruction quality in certain regions, a feature-matching loss can be employed over the activations of certain pre-trained models, \eg, face-embedding network \cite{cao2018vggface2} which can improve the reconstruction quality of face region.

\textbf{Network Architecture.}
Convolution neural network is the common structure to learn the discrete image representation in VQ-VAE.
Recently, Yu \etal \cite{yu2021vector} replace the convolution-based structure with Vision Transformer (ViT) \cite{dosovitskiy2020image}, which is shown to be less constrained by the inductive priors imposed by convolutions and is able to yield better computational efficiency with higher reconstruction quality.
With the emergence of diffusion models, diffusion-based decoder \cite{shi2022divae} also has been explored to learn discrete image representation with superior reconstruction quality.
On the other hand, a multi-scale quantization structure is proved to promote the generation performance by including both low-level pixels and high-level tokens \cite{ni2022n} or hierarchical latent codes \cite{razavi2019generating}.
To further reduce the computational costs, a residual quantization \cite{lee2022autoregressive} can be employed to recursively quantize the image as a stacked map of discrete tokens.

\textbf{Codebook Utilization.}
The vanilla VQ-VAE with argmin operation (to get the nearest codebook entry) suffers from severe codebook collapse, \eg, only few codebook entries are effectively utilized for quantization \cite{zhang2023regularized}.
To alleviate the codebook collapse, vq-wav2vec \cite{baevski2019vq} introduces Gumbel-Softmax \cite{jang2016categorical} to replace argmin for quantization.
The Gumbel-Softmax allows sampling discrete representation in a differentiable way through straight-through gradient estimator \cite{bengio2013estimating}, which boosts the codebook utilization significantly.
ViT-VQGAN \cite{yu2021vector} also presents a factorized code architecture which introduces a linear projection from the encoder output to a low dimensional latent variable space for code index lookup and boosts the codebook usage substantially.

\textbf{Learning Regularization.}
Recent work \cite{shin2021translation} validates that the vanilla VQ-VAE doesn't satisfy translation equivariance during quantization, resulting in degraded performance for text-to-image generation.
A simple but effective TE-VQGAN \cite{shin2021translation} is thus proposed to achieve translation equivariance by regularizing orthogonality in the codebook embeddings.
To regularize the latent structure of heterogeneous domain data in conditional generation,
Zhan \etal \cite{zhan2022auto} design an Integrated Quantization VAE to penalize the inter-domain discrepancy with intra-domain variations.

\begin{figure*}[t]
\centering
\includegraphics[width=1.0\linewidth]{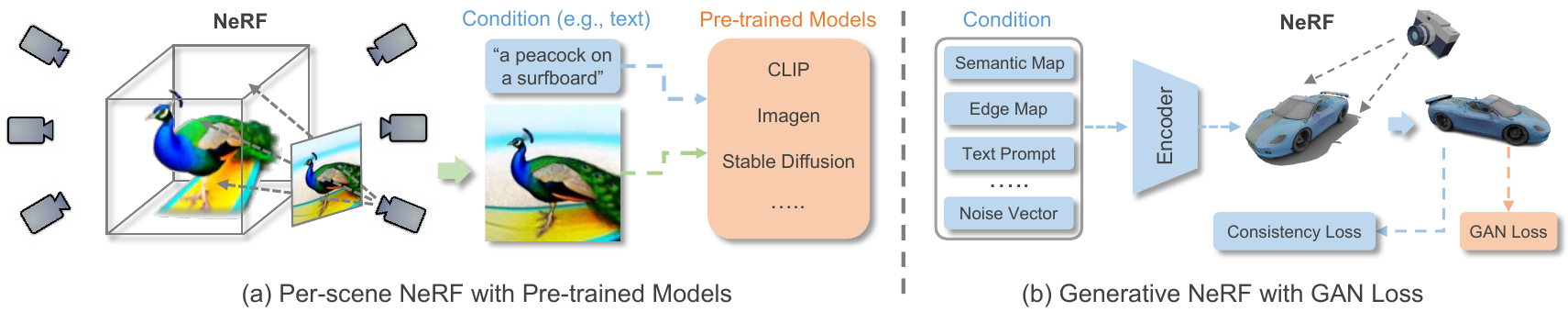}
\caption{
The frameworks of (a) per-scene NeRF and (b) generative (GAN-based) NeRF for 3D-aware MISE. The image is adapted from \cite{poole2023dreamfusion,gao2022get3d}.
}
\label{neural_fields}
\end{figure*}

\subsubsection{Autoregressive Modeling}
Autoregressive (AR) modeling is a representative paradigm to accommodate sequence dependencies, complying with the chain rule of probability.
The probability of each token in the sequence is conditioned on all previously predictions, yielding a joint distribution of sequences as the product of conditional distributions:
$
p(x) = \prod_{t=1}^{n} p(x_{t} | x_1, x_2, \cdots, x_{t-1}) = \prod_{t=1}^{n}p(x_t| x_{<t})
$.
During inference, each token is predicted autoregressively in a raster-scan order. 
Notably, a sliding-window strategy \cite{esser2020taming} can be employed to reduce the cost during inference by only utilizing the predictions within a local window.
A top-$k$ sampling strategy is adopted to randomly sample from the $k$ most likely next tokens, which naturally enables diverse sampling results. The predicted tokens are then concatenated with the previous sequence as conditions for the prediction of next token. This process repeats iteratively until all the tokens are sampled.
Autoregressive models for image synthesis have become increasingly popular due to their ability to generate high-quality, realistic images with a high level of detail. 
In MISE tasks, autoregressive models generate images pixel-by-pixel based on a conditional probability distribution that takes into account both the previously generated pixels and the given conditioning information, which allows the models to capture the complex dependencies to yield visually consistent images.
In recent years, autoregressive models for MISE have been largely fueled by series of designs to be introduced below.

\textbf{Network Architecture.}
Early autoregressive models for image generation usually adopt PixelCNN~\cite{van2016conditional} which struggle in modeling long term relationships within an image due to the limited receptive field.
With the prevailing of Transformer \cite{vaswani2017attention}, 
Transformer-based autoregressive models \cite{parmar2018imagetransformer} emerge with enhanced receptive field which allows sequentially predicting each pixel conditioned on previous prediction results.
To explore the limits of autoregressive text-to-image synthesis, Parti \cite{yu2022scaling} scales the parameter size of Transformer up to 20B, yielding consistent quality improvements in terms of image quality and text-image alignment.
Instead of unidirectionally modeling from condition to image, a bi-directional architecture is also explored in text-to-image synthesis \cite{huang2021picture,huang2021unifying}, which generates both diverse captions and images.

\textbf{Bidirectional Context.}
On the other hand, previous methods incorporate image context in a raster-scan order by attending only to previous generation results.
This strategy is unidirectional and suffers from sequential bias as it disregards much context information until autoregression is nearly complete.
It also ignores much contextual information in different scales as it only processes the image on a single scale.
Grounded in above observations,
ImageBART \cite{esser2021imagebart} presents a coarse-to-fine approach in a unified framework that addresses the unidirectional bias of autoregressive modeling and the corresponding exposure bias.
Specifically, a diffusion process is applied to successively eliminate information, yielding a hierarchy of representations which is further compressed via a multinomial diffusion process \cite{hoogeboom2021argmax,sohl2015deep}.
By modeling the Markovian transition autoregressively with attending to the preceding hierarchical state, crucial global context can be leveraged for each individual autoregressive step.
As an alternative, bidirectional Transformer is also widely explored to incorporate bidirectional context, accompanied with a Masked Visual Token Modeling (MVTM) \cite{chang2022maskgit} or Masked Language Modeling (MLM) \cite{zhang2021m6,yu2021diverse} mechanism.

\textbf{Self-Attention Mechanism.}
To handle languages, images, and videos in different tasks in a unified manner,
NUWA \cite{wu2021nuwa} presents a 3D Transformer framework with a unified 3D Nearby Self-Attention (3DNA) which not only reduces the complexity of full attention but also shows superior performance.
With a focus on semantic image editing at high resolution, ASSET \cite{liu2022asset} proposes to sparsify the Transformer’s attention matrix at high resolutions guided by dense attention at lower resolutions, leading to reduced computational cost.

\subsection{NeRF-based Methods}
\label{nerf}

A neural field \cite{xie2022neural} is a field that is parameterized fully or in part by a neural network.
As a special case of neural fields, Neural Radiance Fields (NeRF) \cite{mildenhall2020nerf} achieve impressive performance for novel views synthesis by parameterizing the color and density of a 3D scene with neural fields.
Specifically, a fully-connected neural network is adopted in NeRF, by taking a spatial location (x, y, z) with the corresponding viewing direction ($\theta$, $\phi$)) as input, and the volume density with the corresponding emitted radiance as output.
To render 2D images from the implicit 3D representation, differentiable volume rendering is performed with a numerical integrator \cite{mildenhall2020nerf} to approximate the intractable volumetric projection integral.  
Powered by NeRF for 3D scene representation, 3D-aware MISE can be achieved with per-scene NeRF or generative NeRF frameworks.

\subsubsection{Per-scene NeRF}
Consistent with the original NeRF model, a per-scene NeRF aims to optimize and represent a single scene supervised by images or certain pre-trained models.

\textbf{Image Supervision.}
With paired guidance and corresponding view images, a NeRF can be naively trained conditioned on the guidance to achieve MISE.
% available head pose information and viewing directions,
For instance, AD-NeRF \cite{guo2021ad} achieves high-fidelity talking-head synthesis by training neural radiance fields on a video sequence with the audio track of one target person.
Instead of bridging audio inputs and video outputs based on the intermediate representations, AD-NeRF directly feeds the audio features into an implicit function to yield a dynamic NeRF, which is further exploited to synthesize high-fidelity talking-face videos accompanied with the audio via volume rendering.
However, the paired condition-image data and multiview images are usually unavailable or costly to acquire which hinders the broad applications of this method.

\textbf{Pre-trained Model Supervision.}
Instead of relying on  multiview images or paired data,
certain pre-trained models can be adopted to optimize NeRFs from scratch as shown in Fig. \ref{neural_fields} (a).
For instance, pre-trained CLIP can be leveraged to achieve text-driven 3D-aware image synthesis \cite{jain2022zero}, by optimizing NeRF to render multi-view images that score highly with a target text description according to the CLIP model.
Similar CLIP-based approach is also adopted in AvatarCLIP \cite{hong2022avatarclip} to achieve zero-shot text-driven 3D avatar generation and animation.
Recently, with the prosperity of diffusion models, pre-trained 2D diffusion models show great potential to drive the generation of high-ﬁdelity 3D scenes for diverse text prompts as in DreamFusion \cite{poole2023dreamfusion}.
Specifically, based on probability density distillation, 2D diffusion model can serve as a generative prior for the optimization of a randomly-initialized 3D neural field via gradient descent such that its 2D renderings yield a high score with the target condition.
Following this line of research, Magic3D \cite{lin2022magic3d} further proposes to optimize a textured 3D mesh model with an efficient differentiable renderer \cite{shen2021deep,gao2022get3d} interacting with a pre-trained latent diffusion model.
On the other hand, optimizing NeRF with pre-trained models is an under-constrained process, which highlights the need of certain prior knowledge or regularizations.
It has been proved that geometric priors including sparsity regularization and scene bounds \cite{jain2022zero} improve the generation fidelity significantly.
Besides, to mitigate the ambiguous geometry from a single viewpoint, random lighting directions can be applied to shade a scene to reveal the geometric details \cite{poole2023dreamfusion}.
To prevent normal vectors from improperly facing backwards from the camera, an orientation loss proposed in Ref-NeRF \cite{verbin2022ref} can be employed to impose penalty.

\subsubsection{Generative NeRF}
Distinct from per-scene optimization NeRFs which work for a single scene,
generative NeRFs are capable of generalizing to different scenes by integrating NeRF with generative models.
In generative NeRF, a scene is specified by a latent code in the corresponding latent space.
GRAF \cite{schwarz2020graf} is the first to introduce a GAN framework for the generative training of radiance fields by employing a multi-scale patch-based discriminator.
Lot of efforts have recently been devoted to improve the generative NeRF, \eg, GIRAFFE \cite{niemeyer2021giraffe} for introducing volume rendering at the feature level and separating the object instances in a controllable way; Pi-GAN \cite{chan2021pi} for the FiLM-based conditioning scheme \cite{perez2018film} with a SIREN architecture \cite{sitzmann2020implicit}; StyleNeRF \cite{gu2021stylenerf} for the integration of style-based generator to achieve high-resolution image synthesis; EG3D \cite{chan2022efficient} for incorporating efficient triplane 3D representation.
Fueled by these advancements, 3D-aware MISE can be well performed following the pipeline of conditional generative NeRF or generative NeRF inversion.

\textbf{Conditional NeRF.}
In conditional generative NeRF, a scene is specified by the combination of 3D positions and given conditions as shown in Fig. \ref{neural_fields} (b).
The condition can be integrated to condition the NeRF following the integration strategies in GANs or diffusion models.
For instance, a pre-trained CLIP model is employed in \cite{jo2021cg} to extract the conditional visual and text features to condition a NeRF.
Similarly, pix2pix3D \cite{deng20233d} encodes certain visual guidance (and a random code) to generate triplanes for scene representation, while it renders the image and pixel-aligned label map simultaneously to enable interactive 3D cross-view editing.

\textbf{NeRF Inversion.}
In the light of recent advances in generative NeRFs for 3D-aware image synthesis, some work explores the inversion of generative NeRFs for 3D-aware MISE.
As generative NeRF (GAN-based) is accompanied with a latent space, the conditional guidance for MISE can be naively mapped into the latent space to enable conditional 3D-aware generation \cite{chen2022sem2nerf}.
However, this method struggles for image generation \& editing with local control.
Some recent work proposes to train 3D-semantic-aware generative NeRF \cite{sun2022ide,sun2022fenerf} that produces spatial-aligned images and semantic masks concurrently with two branches. These aligned semantic masks can be used to perform local editing of 3D volume via NeRF inversion.
On the other hand, the inversion of generative NeRF is challenging due to the including of camera pose. Thus, a hybrid inversion strategy \cite{gu2021stylenerf} can be applied in practice by combining encoder-based and optimization-based inversion, where the encoder predicts a camera pose and a coarse style code which is further refined through inverse optimization.
To enable flexible and faithful 3D-aware MISE, some pre-trained models like CLIP also can be introduced in NeRF inversion.
For instance, to achieve 3D-aware manipulation from text prompt, CLIP-NeRF \cite{wang2021clip} optimizes latent codes towards targeted manipulation driven by a CLIP-based matching loss as described in StyleCLIP \cite{patashnik2021styleclip}.

\renewcommand\arraystretch{1.0}
\begin{table*}[t]
\caption
{
Annotation types in popular datasets for MISE. Note, only currently available annotations are labeled with checkmarks, although some off-the-shelf models (e.g., segmentation models, edge detectors, image caption models) can be employed to annotate the corresponding datasets. Part of the information is retrieved from \cite{xue2022deep}. B-Box denotes bounding box.
}
\renewcommand\tabcolsep{5.5pt}
\centering
\begin{tabular}{l || c c c c c c c c c c }
\hline

Datasets & Samples & Semantic Map & Keypoint & Sketch & B-Box & Depth & Attribute & Text & Audio & Scene Graph \\

\hline
ADE20K \cite{zhou2017ade20k} & 27,574 & \textcolor{green}{\checkmark}  & \textcolor{red}{\xmark}  &  \textcolor{red}{\xmark}   & \textcolor{red}{\xmark} & \textcolor{red}{\xmark} & \textcolor{red}{\xmark}  &  \textcolor{red}{\xmark}   & \textcolor{red}{\xmark} & \textcolor{red}{\xmark}   \\

COCO \cite{lin2014microsoft} & 328,000 & \textcolor{green}{\checkmark}  & \textcolor{green}{\checkmark}  &  \textcolor{red}{\xmark}   & \textcolor{green}{\checkmark} & \textcolor{red}{\xmark} & \textcolor{red}{\xmark}  & \textcolor{green}{\checkmark} & \textcolor{red}{\xmark} & \textcolor{red}{\xmark}  \\

COCO-Stuff \cite{caesar2018cocostuff} & 164,000 & \textcolor{green}{\checkmark}  & \textcolor{red}{\xmark} &  \textcolor{red}{\xmark}   & \textcolor{red}{\xmark} & \textcolor{red}{\xmark} & \textcolor{red}{\xmark}  & \textcolor{red}{\xmark} & \textcolor{red}{\xmark} & \textcolor{green}{\checkmark} \\

PSG \cite{yang2022panoptic} & 48,749 & \textcolor{green}{\checkmark}  & \textcolor{red}{\xmark}  &  \textcolor{red}{\xmark}   & \textcolor{green}{\checkmark} & \textcolor{red}{\xmark} & \textcolor{red}{\xmark}  & \textcolor{green}{\checkmark} & \textcolor{red}{\xmark} & \textcolor{green}{\checkmark}  \\

Cityscapes \cite{cordts2016cityscapes} & 25,000 & \textcolor{green}{\checkmark}  & \textcolor{red}{\xmark}  &  \textcolor{red}{\xmark}   & \textcolor{red}{\xmark} & \textcolor{red}{\xmark} & \textcolor{red}{\xmark}  & \textcolor{red}{\xmark}   & \textcolor{red}{\xmark} & \textcolor{red}{\xmark}   \\

CelebA \cite{liu2015celebahq} & 202,599 & \textcolor{red}{\xmark}  &  \textcolor{green}{\checkmark} &  \textcolor{red}{\xmark}   & \textcolor{red}{\xmark} & \textcolor{red}{\xmark} &  \textcolor{green}{\checkmark} & \textcolor{red}{\xmark}  & \textcolor{red}{\xmark} & \textcolor{red}{\xmark}   \\

CelebA-HQ \cite{karras2017progressive} & 30,000 & \textcolor{red}{\xmark}  &  \textcolor{green}{\checkmark} &  \textcolor{red}{\xmark}   & \textcolor{red}{\xmark} & \textcolor{red}{\xmark} &  \textcolor{green}{\checkmark} & \textcolor{red}{\xmark}  & \textcolor{red}{\xmark} & \textcolor{red}{\xmark}   \\

CelebAMask-HQ \cite{lee2020maskgan} & 30,000 & \textcolor{green}{\checkmark}  & \textcolor{green}{\checkmark}  &  \textcolor{red}{\xmark}   & \textcolor{red}{\xmark} & \textcolor{red}{\xmark} &  \textcolor{green}{\checkmark}   & \textcolor{red}{\xmark} & \textcolor{red}{\xmark} & \textcolor{red}{\xmark}   \\

CelebA-Dialog \cite{jiang2021talk} & 202,599 & \textcolor{red}{\xmark}  & \textcolor{red}{\xmark}  &  \textcolor{red}{\xmark}   & \textcolor{red}{\xmark} & \textcolor{red}{\xmark} &  \textcolor{green}{\checkmark}   & \textcolor{green}{\checkmark} & \textcolor{red}{\xmark} & \textcolor{red}{\xmark}   \\

MM-CelebA-HQ \cite{xia2021tedigan} & 30,000 & \textcolor{green}{\checkmark} &  \textcolor{green}{\checkmark} &  \textcolor{green}{\checkmark}  & \textcolor{red}{\xmark} & \textcolor{red}{\xmark} &  \textcolor{green}{\checkmark} & \textcolor{green}{\checkmark}  & \textcolor{red}{\xmark} & \textcolor{red}{\xmark}   \\

DeepFashion \cite{liu2016deepfashion} & 800,000 & \textcolor{red}{\xmark}   & \textcolor{green}{\checkmark} &  \textcolor{red}{\xmark}   & \textcolor{red}{\xmark} & \textcolor{red}{\xmark} &  \textcolor{green}{\checkmark}  & \textcolor{red}{\xmark} & \textcolor{red}{\xmark} & \textcolor{red}{\xmark}   \\

DeepFashion-MM \cite{jiang2022text2human} & 44,096 &  \textcolor{green}{\checkmark}  & \textcolor{green}{\checkmark} &  \textcolor{red}{\xmark}   & \textcolor{red}{\xmark} & \textcolor{red}{\xmark} &  \textcolor{green}{\checkmark}  &  \textcolor{green}{\checkmark} & \textcolor{red}{\xmark} & \textcolor{red}{\xmark}   \\

Chictopia10K \cite{liang2015deep} & 14,400 & \textcolor{green}{\checkmark}  & \textcolor{red}{\xmark}  &  \textcolor{red}{\xmark}   & \textcolor{red}{\xmark} & \textcolor{red}{\xmark} & \textcolor{red}{\xmark}  &  \textcolor{red}{\xmark}   & \textcolor{red}{\xmark} & \textcolor{red}{\xmark}   \\

NYU Depth \cite{silberman2011indoor} & 1,449 & \textcolor{green}{\checkmark}  & \textcolor{red}{\xmark}  &  \textcolor{red}{\xmark}   & \textcolor{red}{\xmark} & \textcolor{green}{\checkmark} & \textcolor{red}{\xmark}  &  \textcolor{red}{\xmark}   & \textcolor{red}{\xmark} & \textcolor{red}{\xmark}   \\

Stanford’s Cars \cite{krause20133d} & 16,185 & \textcolor{red}{\xmark} & \textcolor{red}{\xmark}  &  \textcolor{red}{\xmark}   & \textcolor{green}{\checkmark} & \textcolor{green}{\checkmark} & \textcolor{green}{\checkmark}  &  \textcolor{red}{\xmark}   & \textcolor{red}{\xmark} & \textcolor{red}{\xmark}   \\

Oxford-102 \cite{nilsback2008automated} & 8,189 & \textcolor{red}{\xmark}  & \textcolor{red}{\xmark}  &  \textcolor{red}{\xmark}   & \textcolor{red}{\xmark} & \textcolor{red}{\xmark} & \textcolor{green}{\checkmark}  &  \textcolor{green}{\checkmark}   & \textcolor{red}{\xmark} & \textcolor{red}{\xmark} \\

CUB-200 \cite{welinder2010caltech} & 11,788 & \textcolor{green}{\checkmark} & \textcolor{red}{\xmark}  &  \textcolor{red}{\xmark}   & \textcolor{green}{\checkmark} & \textcolor{red}{\xmark} & \textcolor{green}{\checkmark}  &  \textcolor{green}{\checkmark}   & \textcolor{red}{\xmark} & \textcolor{red}{\xmark} \\

LAION-5B \cite{schuhmannlaion} & 5,85 billion & \textcolor{red}{\xmark} & \textcolor{red}{\xmark}  &  \textcolor{red}{\xmark}   & \textcolor{red}{\xmark} & \textcolor{red}{\xmark} & \textcolor{red}{\xmark}  &  \textcolor{green}{\checkmark}   & \textcolor{red}{\xmark} & \textcolor{red}{\xmark} \\

Visual Genome \cite{krishna2017visual} & 101,174 & \textcolor{red}{\xmark}  & \textcolor{red}{\xmark}  &  \textcolor{red}{\xmark}   & \textcolor{green}{\checkmark} & \textcolor{red}{\xmark} & \textcolor{green}{\checkmark}  & \textcolor{green}{\checkmark}  & \textcolor{red}{\xmark} & \textcolor{green}{\checkmark} \\

VoxCeleb \cite{nagrani2017voxceleb} & 148,642 & \textcolor{red}{\xmark}  & \textcolor{red}{\xmark}  &  \textcolor{red}{\xmark}   & \textcolor{red}{\xmark} & \textcolor{red}{\xmark} & \textcolor{red}{\xmark}  &  \textcolor{red}{\xmark}   & \textcolor{green}{\checkmark} & \textcolor{red}{\xmark}   \\

LRS \cite{son2017lip} & 144,482 & \textcolor{red}{\xmark}  & \textcolor{red}{\xmark}  &  \textcolor{red}{\xmark}   & \textcolor{red}{\xmark} & \textcolor{red}{\xmark} & \textcolor{red}{\xmark}  &  \textcolor{green}{\checkmark}   & \textcolor{green}{\checkmark} & \textcolor{red}{\xmark}   \\

\hline
\end{tabular}
\label{tab_datasets}
\end{table*}

\subsection{Other Methods}
\label{others}

Except for above-mentioned methods, there has been several endeavors dedicated to the MISE task, exploring diverse research paths.

\textbf{2D MISE without Generative Models.}
Instead of relying on generative models, a series of alternative methods have been explored for multimodal editing of 2D images.
For instance, CLVA \cite{fu2021language} manipulates the style of a content image through text prompts by comparing the contrastive pairs of content image and style instruction to achieve mutual relativeness.
However, CLVA is constrained as it requires style images accompanied with the text prompts during training.
Instead, CLIPstyler \cite{kwon2021clipstyler} leverages pre-trained CLIP model to achieves text guided style transfer by training a lightweight network which transforms a content image to follow the text condition.
As an extension to video, Loeschcke \etal \cite{loeschcke2022text} harness the power of CLIP to stylize the object in a video according to two target texts.

\textbf{3D-aware MISE without NeRF.}
Except for NeRF, there are alternative methods that can be leveraged for 3D-aware MISE.
Typically, classical 3D representations such as mesh also can be employed to replace NeRF for 3D-aware MISE \cite{michel2022text2mesh,khalid2022text}.
Specifically, aiming for style transfer of 3D scenes, Mu \etal \cite{mu20223d} propose to learn geometry-aware content features from a point cloud representation of the scene, followed by point-to-pixel adaptive attention normalization (AdaAttN) to transfer the style of a given image.
Besides, a popular line of research adapts GANs for 3D-aware generation by conditioning on camera parameters \cite{noguchi2019rgbd}, introducing intermediate 3D shape \cite{zhu2018visual}, incorporating depth prior \cite{shi20223d}, and adopting 3D rigid-body transformation with projection \cite{nguyen2019hologan}.

\subsection{Comparison and Discussion}

All generation methods possess their own strength and weakness. 
GAN-based methods can achieve high-fidelity image synthesis in terms of FID and Inception Score and also have fast inference speed, while GANs are notorious for unstable training and are prone to mode collapse. 
Moreover, it has been shown that GANs focus more on fidelity rather than capturing the diversity of the training data distribution compared with likelihood-based models like diffusion models and autoregressive models \cite{dhariwal2021diffusion}.
Besides, GANs usually adopt a CNN architecture (although Transformer structure is explored in some studies \cite{jiang2021transgan,park2022styleformer,hudson2021generative}), which makes them struggle to handle multimodal data in a unified manner and generalize to new MISE tasks.
With wide adoption of Transformer backbone, autoregressive models can handle different MISE tasks in a unified manner.
However, thanks to the autoregressive prediction of tokens, autoregressive models suffer from slow inference speed, which is also a bottleneck of diffusion models as requiring a number of diffusion steps.
Currently, autoregressive models and diffusion models are more favored in SOTA methods compared with GANs, especially for text-to-image synthesis.

Autoregressive models and diffusion models are likely-based generative models which are equipped with stationary training objective and good training stability.
The comparison of generative modeling capability between autoregressive models and diffusion is still inconclusive.
DALL-E 2 \cite{ramesh2022hierarchical} shows that diffusion models are slightly better than autoregressive models in modeling the diffusion prior.
However, the recent work Parti \cite{yu2022scaling} which adopts an autoregressive structure presents superior performance over the SOTA work of diffusion-based methods (i.e., Imagen).
On the other hand, the exploration of two different families of generative models may open exciting opportunities to combine the merits of the two powerful models.

Different from above generation methods which mainly work on 2D images and have few requirements for the training datasets, NeRF-based methods handle the 3D scene geometry and thus have relatively high requirements for training data. For example, per-scene optimization NeRFs require multiview images or video sequence with pose annotation, while generative NeRFs require the scene geometry of the dataset to be simple.
Thus, the application of NeRF in MISE with high-fidelity is still quite constrained.
Nevertheless, the 3D-aware modeling of real world with NeRF opens a new door for future MISE research, broadening the horizons for potential advancements.

Besides, state-of-the-art methods are prone to combine different generative models to yield superior performance. For example, Taming Transformer \cite{esser2020taming} incorporates VQ-GAN and Autoregressive models to achieve high-resolution image synthesis; StyleNeRF \cite{gu2021stylenerf} combines NeRF with GAN to enable high-fidelity image synthesis with both high-fidelity and 3D-awareness; ImageBart \cite{esser2021imagebart} combines the autoregressive formulation with a multinomial diffusion process to incorporate a coarse-to-ﬁne hierarchy of context information;
X-LXMERT \cite{cho2020x} integrates GAN into the framework of cross-modality representation to achieve text-guided image generation.

\begin{figure*}[t]
\centering
\includegraphics[width=1.0\linewidth]{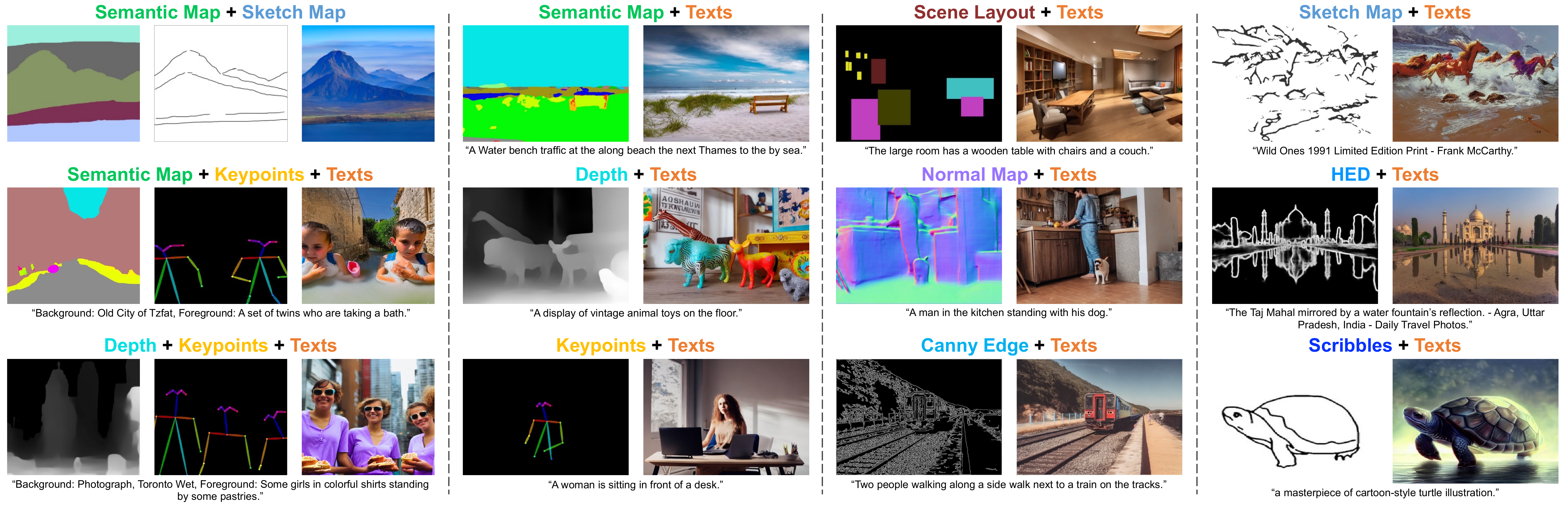}
\caption{
Image synthesis from the combination of different types of guidance.
The samples are from \cite{qin2023unicontrol,zhang2023adding,huang2021multimodal}.
}
\label{im_gallery}
\end{figure*}

\renewcommand\arraystretch{0.9}
\begin{table*}[t]
    \caption{Quantitative comparison with existing methods on segmentation-to-image synthesis. Part of the results are retrieved from \cite{wang2022semantic}.
    }
    \renewcommand\tabcolsep{5.9pt}
    \centering
    \begin{tabular}{l ccc ccc ccc ccc}
    \toprule
     & \multicolumn{3}{c}{CelebAMask-HQ} & \multicolumn{3}{c}{Cityscapes} & \multicolumn{3}{c}{ADE20K} & \multicolumn{3}{c}{COCO-Stuff} \\
    \cmidrule(r){2-4} \cmidrule(r){5-7} \cmidrule(r){8-10} \cmidrule(r){11-13} 
    \multirow{-2}{*}{Methods} & FID $\downarrow$ & LPIPS $\uparrow$ & mIoU $\uparrow$ & FID $\downarrow$ & LPIPS $\uparrow$ & mIoU $\uparrow$ & FID $\downarrow$ & LPIPS $\uparrow$ & mIoU $\uparrow$ & FID $\downarrow$ & LPIPS $\uparrow$ & mIoU $\uparrow$ \\
    \midrule
    {Pix2PixHD~\cite{wang2018high}} & 38.5 & - & 76.1     & 95.0 & - & 63.0      & 81.8 & - & 28.8      & 111.5 & - & 26.6  \\
    {SPADE~\cite{park2019semantic}} & 29.2 & - & 75.2  & 71.8 & - & 61.2     & 22.6 & - & 38.3   & 33.9 & - & 38.4  \\
    {CLADE~\cite{tan2021efficient}} & 30.6 & - & 75.4     & 57.2 & - & 58.6     & 35.4 & -  & 23.9     & 29.2 & - & 38.8 \\ 
    {CC-FPSE~\cite{liu2019learning}} & - & - & -       & 54.3 & 0.026 & 65.2     & 31.7 & 0.078 & 40.6   & 19.2 & 0.098 & 42.9 \\
    {GroupDNet~\cite{zhu2020semantically}} & 25.9 & 0.365 & 76.1    & 47.3 & 0.101 & 55.3     & 41.7 & 0.230  & 27.6   &  - & - & -   \\
    {INADE~\cite{tan2021diverse}} & 21.5 & 0.415 & 74.1     & 44.3 & 0.295 & 57.7        & 35.2 & 0.459 & 33.0      & - & - & -   \\
    {OASIS~\cite{sushko2020you}} & - & - & - & 47.7 & 0.327 & 58.3 & 28.3 & 0.286 & \textbf{45.7} & 17.0 & 0.328 & \textbf{46.7}  \\ 
    \rowcolor{mygray} {Taming \cite{esser2020taming}} & - & - & - & - & - & - & 35.5 & 0.421 & - & - & - & - \\
    \rowcolor{LightCyan} {SDM~\cite{wang2022semantic}} & \textbf{18.8} & \textbf{0.422} & \textbf{77.0}    & \textbf{42.1} & \textbf{0.362} & \textbf{77.5} & \textbf{27.5} & \textbf{0.524} & 39.2 & \textbf{15.9} & \textbf{0.518} & 40.2 \\
    \bottomrule
    \end{tabular}
    % \vspace{-2mm}
    % \vspace{-6mm}
    \label{tab_visual}
\end{table*}

\setlength{\tabcolsep}{10pt}
\begin{table}[tbh]
\small
\begin{center}
\caption{
Text-to-Image generation performance on the COCO dataset.
\textsuperscript{\textdagger} denotes the results obtained by using the corresponding open-source code. The rows in grey and cyan denote the results of Transformer-based and Diffusion-based methods, respectively. Others are the results of GAN-based methods. Part of the results are retrieved from \cite{frolov2021adversarial}.
}
\begin{tabular}{lrrr}
        \toprule
        Methods         &  IS $\uparrow$ & FID $\downarrow$ & R-Prec. $\uparrow$        \\
        \midrule
        Real Images \cite{hinz2019semantic} & 34.88  & 6.09 & 68.58 \\
        \midrule
        StackGAN \cite{zhang2017stackgan} & 8.450 & 74.05 & - \\
        StackGAN++ \cite{zhang2018stackgan++} & 8.300 & 81.59 & - \\
        AttnGAN \cite{xu2018attngan} & 25.89 & 35.20 & 85.47 \\
        MirrorGAN \cite{qiao2019mirrorgan} & 26.47 & - & 74.52 \\
        AttnGAN+OP \cite{hinz2019semantic} & 24.76 & 33.35 & 82.44 \\
        OP-GAN \cite{hinz2019semantic}& 27.88 & 24.70 & 89.01  \\
        SEGAN \cite{tan2019semantics} & 27.86 & 32.28 & - \\
        ControlGAN \cite{li2019controllable} & 24.06 & - & 82.43 \\
        DM-GAN \cite{zhu2019dm} & 30.49 & 32.64 & 88.56  \\
        DM-GAN \cite{zhu2019dm}\textsuperscript{\textdagger} & 32.43 & 24.24 & 92.23 \\
        Obj-GAN \cite{li2019object} & 27.37 & 25.64 & 91.05 \\
        Obj-GAN \cite{li2019object}\textsuperscript{\textdagger} & 27.32 & 24.70 & 91.91 \\
        TVBi-GAN \cite{wang2020text} & 31.01 & 31.97 & - \\
        Wang et al. \cite{wang2020end} & 29.03  & 16.28 & 82.70 \\
        Rombach et al. \cite{rombach2020network} & 34.70 & 30.63 & - \\
        CPGAN \cite{liang2020cpgan} & \textbf{52.73} & - & \textbf{93.59} \\
        Pavllo et al. \cite{pavllo2020controlling} & - & 19.65 & - \\
        XMC-GAN \cite{zhang2021cross} & 30.45 & 9.330 & - \\
        LAFITE \cite{zhou2021lafite} & 32.34 & 8.120 & - \\
\rowcolor{mygray} CogView \cite{ding2021cogview} & 18.20 & 27.10 & - \\
\rowcolor{mygray} CogView2 \cite{ding2022cogview2} & 22.40 & 24.10 & - \\
\rowcolor{mygray} DALL-E \cite{ramesh2021dalle} & 17.90 & 27.50 & - \\
\rowcolor{mygray} NUWA \cite{wu2021nuwa} & 27.20 & 12.90 & - \\
\rowcolor{mygray} DiVAE \cite{shi2022divae} & - & 11.53 & - \\
\rowcolor{mygray} Make-A-Scene \cite{gafni2022make} & - & 11.84 & - \\
\rowcolor{mygray} Parti \cite{yu2022scaling} & - & \textbf{7.230} & - \\
\rowcolor{LightCyan} VQ-Diffusion \cite{gu2021vector} & - & 13.86 & - \\
\rowcolor{LightCyan} LDM \cite{rombach2022high} & 30.29 & 12.63 & - \\
\rowcolor{LightCyan} GLIDE \cite{nichol2021glide} & - & 12.24 & - \\
\rowcolor{LightCyan} DALL-E 2 \cite{ramesh2022hierarchical} & - & 10.39 & - \\
\rowcolor{LightCyan} Imagen \cite{saharia2022photorealistic} & - & 7.270 & - \\
        \bottomrule
    \end{tabular}
    \label{tab_coco}
\end{center}
\end{table}

\section{Experimental Evaluation}
\label{experiments}

\subsection{Datasets}
Datasets are the core of image synthesis and editing tasks.
To give an overall picture of the datasets in MISE, we tabulate the detailed annotation types in popular datasets in Table \ref{tab_datasets}. 
Notably, ADE20K \cite{zhou2017ade20k}, COCO-Stuff \cite{caesar2018cocostuff}, and Cityscapes \cite{cordts2016cityscapes} are common benchmark datasets for semantic image synthesis; Oxford-120 Flowers \cite{nilsback2008automated}, CUB-200 Birds \cite{welinder2010caltech}, and COCO \cite{lin2014microsoft} are widely adopted in text-to-image synthesis; VoxCeleb2 \cite{chung2018voxceleb2} and Lip Reading in the Wild (LRW) \cite{chung2016lip} are usually used for the benchmark of taking face generation.
Please refer to the supplementary material for more details of the widely adopted datasets in different MISE tasks.

\vspace{5pt}

\subsection{Evaluation Metrics}
Precise evaluation metrics are of great importance in driving progress of research. On the other hand, the evaluation of MISE tasks is challenging as multiple attributes account for a fine generation result and the notion of image evaluation is often subjective.
To achieve faithful evaluation, comprehensive metrics are adopted to evaluate MISE tasks from multiple aspects.
Specifically, Inception Score (IS) \cite{salimans2016improved} and FID \cite{fid} are general metrics for image quality evaluation, while LPIPS \cite{zhang2018unreasonable} is a common metric to evaluate image diversity.
These metrics can be applied across different generation tasks.
In terms of the alignment between generated images and conditions, the evaluation metrics are usually designed for specific generation tasks, e.g., mIoU and mAP for semantic image synthesis, 
R-precision \cite{xu2018attngan}, Captioning Metrics \cite{hong2018inferring} and Semantic Object Accuracy (SOA) \cite{hinz2019semantic} for text-to-image generation, 
Landmark distance (LMD) and audio-lip synchronization (Sync) \cite{chung2016out} for talking face generation.

As a general image quality metric, the advantage of IS is its simplicity, and it can be applied to a wide range of image generation models. 
However, IS has been criticized for its lack of robustness and sensitivity to noise.
It also struggles to evaluate overfitting generation (i.e., the model memorizes the training set) and measure intra-domain variation (i.e., the model only produces one good sample).
FID is more robust than the IS and can better capture the overall quality of the generated images. However, it assumes a Gaussian distribution for image features which is not always valid.
For diversity evaluation metrics like LPIPS, the quality of generated images is not concerned which means unrealistic generation could lead to a good diversity score.
Alignment metric provides quantitative evaluations of generation alignment, while most of them are subject to various issues, including insensitivity to temporal or overall coherence in SOA and CPBD, dataset or pre-trained model bias in R-precision, mIoU \& mAP and audio-lip synchronization, ambiguous alignment in Captioning Metrics.
Please refer to the supplementary material for more details of the corresponding evaluation metrics.
Overall, certain evaluation metric should be applied in conjunction with other evaluation metrics for a comprehensive and faithful analysis of model performance.

\subsection{Experimental Results}

To showcase the capability and effectiveness of MISE in a tangible manner, we visualize the synthesized images conditioned on the combination of diverse guidance types as shown in Fig. \ref{im_gallery}. Please refer to the supplementary material for more visualization.
Furthermore, we provide a quantitative comparison of the image synthesis performance exhibited by various models. This assessment takes into consideration distinct types of guidance including visual, text, and audio, which will be discussed in the following sections.

\subsubsection{Visual Guidance}
For visual guidance, we mainly conduct comparison on semantic image synthesis as there are numbers of methods for benchmarking.
As shown in Table \ref{tab_visual}, the experimental comparison is conducted on four challenging datasets: ADE20K \cite{zhou2017ade20k}, ADE20K-outdoors \cite{zhou2017ade20k}, COCO-stuff \cite{caesar2018cocostuff} and Cityscapes \cite{cordts2016cityscapes}, following the setting of \cite{park2019semantic}. 
The evaluation is performed with FID, LPIPS, and mIoU.
Specially, the mIoU aims to assess the alignment between the generated image and the ground truth segmentation via a pre-trained semantic segmentation network.
Pre-trained UperNet101 \cite{xiao2018unified}, multi-scale DRN-D-105 \cite{yu2017dilated}, and DeepLabV2 \cite{chen2014semantic} are adopted for Cityscapes, ADE20K \& ADE20K-outdoors, and COCO-Stuff, respectively.

As shown in Table \ref{tab_visual}, diffusion-based method (i.e., SDM \cite{wang2022semantic}) achieves superior generation quality and diversity as evaluated by FID and LPIPS, and yields comparable semantic consistency as evaluated by mIoU compared with GAN-based methods.
Although the comparison may not be fair as the model sizes are different, diffusion-based method still demonstrates its powerful modeling capability for semantic image synthesis.
With a large model size, autoregressive method Taming \cite{esser2020taming} doesn't show a clear advantage over other methods.
We conjecture that Taming Transformer \cite{esser2020taming} is a versatile framework for various conditional generation tasks without specific design for semantic image synthesis, while other methods in Table \ref{tab_visual} mainly focus on the task of semantic image synthesis.
Notably, autoregressive method and diffusion method inherently support diverse conditional generation results, while GAN-based methods usually require additional modules (\eg, VAE \cite{kingma2013auto}) or designs to achieve diverse generation.

\subsubsection{Text Guidance}
We benchmark text-to-image generation methods on COCO dataset as tabulated in Table \ref{tab_coco} (The results are extracted from relevant papers).
As shown in Table \ref{tab_coco}, GAN-based, autoregressive, and diffusion-based methods can all achieve SOTA performance in terms of FID, e.g., 8.12 in GAN-based method LAFITE \cite{zhou2021lafite}, 7.23 in autoregressive method Parti \cite{yu2022scaling}, and 7.27 in diffusion-based method Imagen \cite{saharia2022photorealistic}.
However, autoregressive and diffusion-based methods are still preferred in recent SOTA work, thanks to their stationary training objective and good scalability \cite{dhariwal2021diffusion}.

\renewcommand\arraystretch{0.9}
\begin{table}[t]
\caption{
The audio guided image editing (talking-head) performance on LRW~\cite{chung2016lip} and VoxCeleb2~\cite{chung2018voxceleb2} under three metrics.
% $^{\dagger}$ denotes that the model is evaluated by directly using the authors' generated samples under their setting. 
$^{\star}$ denotes that the model is trained for subject-specific talking-head generation. Part of the results are retrieved from \cite{zhou2021pose}.
}
\renewcommand\tabcolsep{1.5pt}
\centering
% \vspace{-2pt}
\begin{tabular}{lcccccccc}
\toprule
% \hline
 & \multicolumn{4}{c}{LRW~\cite{chung2016lip}} & \multicolumn{4}{c}{VoxCeleb2~\cite{chung2018voxceleb2}} \\
\cmidrule(lr){2-4} \cmidrule(lr){5-7}
% \cline{2-9}
\multirow{-2}{*}{Methods} & SSIM $\uparrow$ & LMD $\downarrow$ & Sync $\uparrow$ & SSIM $\uparrow$  &  LMD $\downarrow$  &  Sync $\uparrow $\\
% \hline
\midrule  
ATVG~\cite{chen2019hierarchical}  &0.810  & 5.25  &4.1 &0.826 &6.49 &4.3 \\
Wav2Lip~\cite{prajwal2020lip}   &0.862 & 5.73 &\textbf{6.9} &0.846 &12.26 &4.5 \\
MakeitTalk~\cite{zhou2020makelttalk}  &0.796  & 7.13&3.1 &0.817 &31.44 &2.8\\
Rhythmic Head~\cite{chen2020talking} &-  & - &- &0.779 &14.76 &3.8 \\
PC-AVS \cite{zhou2021pose} & 0.861 &3.93 &6.4 &\textbf{0.886} & 6.88 &5.9 \\
GC-AVT \cite{liang2022expressive} &-  & - &- &0.710 &3.03 &5.3 \\
EAMM \cite{ji2022eamm} &0.740  &2.08  &5.5 &- &- &- \\
SyncTalkFace \cite{park2022synctalkface} &\textbf{0.893} & \textbf{1.25}  &- &- &- &- \\
DIRFA \cite{wu2023audio} & -  & 3.16  & 6.4 &- &4.45 &5.8 \\
AVCT$^{\star}$ \cite{wang2022one} &-  & -  &- &- & \textbf{0.25} & \textbf{7.0} \\
\rowcolor{orange} Ground Truth & 1.000 & 0.00 & 6.5 & 1.000 &0.00 &  5.9 \\
\bottomrule
\end{tabular}
% \end{center}
\label{tab_talking_face}
% \vspace{-12pt}
\end{table}

\subsubsection{Audio Guidance}
In terms of audio guided image synthesis and editing, we conduct quantitative comparison in the task of audio-driven talking face generation which has been widely explored in the literature.
Notably, current development of talking face generation mainly relies on GANs, while autoregressive or diffusion-based methods for talking face generation remain under-explored.
The quantitative results of talking face generation on LRW \cite{chung2016lip} and VoxCeleb2 \cite{chung2018voxceleb2} datasets are shown in Table \ref{tab_talking_face}.

\section{Open Challenges \& Discussion}
\label{future}

Though MISE has made notable progress and achieved superior performance in recent years,
there exist several challenges for future exploration.
In this section, we overview the typical challenges, share our humble opinions on possible solutions, and highlight the future research directions.

\subsection{Towards Large-Scale Multi-Modality Datasets}

As current datasets mainly provide annotations in a single modality (\eg, visual guidance), most existing methods focus on image synthesis and editing conditioned on guidance from a single modality (\eg, text-to-image synthesis, semantic image synthesis).
However, humans possess the capability of creating visual contents with guidance of multiple modalities concurrently.
Targeting to mimic the human intelligence, multimodal inputs are expected to be fused and leveraged jointly in image generation.
Recently, Make-A-Scene \cite{gafni2022make} explores to include semantic segmentation tokens in autoregressive modeling to achieve better quality in image synthesis;
ControlNet \cite{zhang2023adding} incorporates various visual conditions into Stable Diffusion (for text-to-image generation) to achieve controllable generation;
with MM-CelebA-HQ \cite{xia2021tedigan}, COCO \cite{lin2014microsoft}, and COCO-Stuff \cite{caesar2018cocostuff} as the training set,
PoE-GAN \cite{huang2021multimodal} achieves image generation conditioned on multi-modal including segmentation, sketch, image, and text. 
However, the size of MM-CelebA-HQ \cite{xia2021tedigan}, COCO \cite{lin2014microsoft}, and COCO-Stuff \cite{caesar2018cocostuff} is still far from narrowing the gap with real-world distributions.
Therefore, to encompass a broad range of modalities into image generation, there is a need for a large-scale dataset equipped with annotations spanning a wide spectrum of modalities, such as semantic segmentation, text descriptions, and scene graphs.
One potential approach to assemble such a dataset could be utilizing pre-trained models for different tasks to generate the requisite annotations. For instance, a segmentation model could be used to create semantic maps, a detection model could be employed to annotate bounding boxes.
Additionally, synthetic data could provide another feasible alternative, given its inherent advantage of readily providing a multitude of annotations.

\subsection{Towards Faithful Evaluation Metrics}
Accurate yet faithful evaluation is of great significance for the development of MISE and is still an open problem.
Leveraging pre-trained models to conduct evaluations (\eg, FID) is constrained to the pre-trained datasets, which tends to pose discrepancy with the target datasets.
User study recruits human subjects to assess the synthesized images directly, which is however often resource-intensive in terms of time and cost.

With the advance of multimodal pre-training, CLIP \cite{radford2021learning} is used to measure the similarity between the texts and generated images, which however does not correlate well with human preferences.
To inherit the merits of powerful representation of pre-trained models and human preference of crowd-sourcing study, fine-tuning pre-trained CLIP with human preference datasets \cite{wu2023human,kirstain2023pick} will be a promising direction for the designing of MISE evaluation metrics.

\subsection{Towards Efficient Network Architecture}
With inherent support for multimodal input and powerful generative modeling, autoregressive models and diffusion models have been a new paradigm for unified MISE.
However, both autoregressive models and diffusion models suffer from slow inference speed, which is more severe in high-resolution image synthesis.
Some works \cite{jayaram2021parallel,dockhorn2021score} explore to accelerate autoregressive models and diffusion models, while the experiments are constrained to toy datasets with low resolution.
Recently, Song et al. \cite{song2023consistency} introduced consistency models based on diffusion processes, which allow to generate high quality samples by directly mapping noise to data with a support of fast one-step generation and multistep sampling to trade compute for sample quality.
The sampling efficiency of this model architecture presents a compelling opportunity for the advancements of network architecture in MISE tasks.

\subsection{Towards 3D Awareness}
With the emergence of neural scene representation models especially NeRF, 3D-aware image synthesis and editing has the potential to be the next breaking point for MISE as it models the 3D geometry of real world.
With the incorporation of generative models, generative NeRF is notably appealing for MISE as it is associated with a latent space.
Current generative NeRF models (\eg, StyleNeRF, EG3D) have enabled to model scenes with simple geometry (\eg, faces, cars) from a collection of unposed 2D images, just like the training of unconditional GANs (\eg, StyleGAN).
Powered by these efforts, several 3D-aware MISE tasks have been explored, \eg, text-to-NeRF \cite{jo2021cg} and semantic-to-NeRF \cite{chen2022sem2nerf}.
However, current generative NeRFs still struggle on datasets with complex geometry variation, \eg, DeepFashion \cite{liu2016deepfashion} and ImageNet \cite{deng2009imagenet}.

Only relying on generative models to learn the complex scene geometry from unposed 2D images is indeed intractable and challenging.
A possible solution is to provide more prior knowledge of the scene, \eg, obtaining prior geometry with off-the-shelf models \cite{skorokhodov3d}, providing skeleton prior for generative human modeling, etc.
Notably, the power of prior knowledge has been explored in some recent studies of 3D-aware tasks \cite{skorokhodov3d,mu20223d,xu2022point}.
Another possible approach is to provide more supervision, \eg, creating a large dataset with multiview annotations or geometry information.
Once the 3D-aware generative modeling succeeds to work on complex natural scenes, some interesting multimodal applications will become possible, \eg, 3D version of DALL-E.

\section{Social Impacts}
\label{social}
As related to the hot concept of \textbf{AI}-\textbf{G}enerated \textbf{C}ontent (\textbf{AIGC}), MISE has gained considerable attention in recent years.
The rapid advancements in MISE offer unprecedented generation realism and editing possibilities, which have influenced and will continue to influence our society in both positive and potentially negative ways.
In this section, we discuss the correlation between MISE and AIGC, and analyze the potential social impacts of MISE.

\subsection{Correlation with AIGC}
Recently, AI-generated content has been a very hot research topic with emergence of Stable Diffusion and ChatGPT.
MISE is related to AIGC in that they both involve using machine learning \& deep learning to create new and novel visual contents.
Nevertheless, MISE is a specific application of AI that focuses on generating \& editing images with specific attributes which is controlled by various multimodal guidance.
It aims to mimic the visual imaging capability of humans in the multimodal real world.
As a comparison, AIGC encompasses a much broader range of creative work including visual contents, text contents, audio contents, etc.

\subsection{Applications}
The multi-modal image synthesis and editing techniques can be applied in artistic creation and content generation, which could widely benefit designers, photographers, and content creators \cite{bailey2020tools}. Moreover, they can be democratized in everyday applications as image generation or editing tools for popular entertainment. 
In addition, the various conditions as intermediate representations for synthesis \& editing greatly ease the use of the methods and improve the flexibility of user interaction. 
In general, the techniques greatly lower the barrier for the public and unleash their creativity on content generation and editing.

\subsection{Misuse}
On the other hand, the increasing editing capability and generation realism also offers opportunities to generate or manipulate images for malicious purposes. The misuse of synthesis \& editing techniques may spread fake or nefarious information and lead to negative social impacts. 
To prevent potential misuses, one possible way is to develop detection techniques for automatically identifying generated images, which has been actively researched by the community \cite{mirsky2021creation}. Meanwhile, sufficient guardrails, labelling, and access control should be carefully considered when deploying MISE techniques to minimize the risk of misuses.

\subsection{Environment} 
As deep-learning-based methods, the current multi-model generative methods inevitably require GPUs and considerable energy consumption for training and inference, which may negatively influence the environment and global climate before the large-scale use of renewable energy. One direction to soften the need for computational resources lies in the active exploration of model generalization. For example, a pretrained model generalized in various datasets could greatly accelerate the training process or provide semantical knowledge for downstream tasks.

\section{Conclusion}
\label{conclusion}

This review has covered main approaches for multimodal image synthesis and editing.
Specifically, we provide an overview of different guidance modalities including visual guidance, text guidance, audio guidance, and other modality guidance (\eg, scene graph).
In addition, we provided a detailed introduction of the main image synthesis \& editing paradigms: GAN-based methods, diffusion-based methods, autoregressive methods, and NeRF-based methods.
The corresponding strengths and weaknesses were comprehensively discussed to inspire new paradigm that takes advantage of the strengths of existing frameworks.
We also conduct a comprehensive survey of datasets and evaluation metrics for MISE conditioned on different guidance modalities. 
Further, we tabularize and compare the performance of existing approaches in different MISE tasks.
Last but not least, we provided our perspective on the current challenges and future directions related to integrating all modalities, comprehensive datasets, evaluation metrics, model architecture, and 3D awareness.

\section*{Acknowledgments}
Fangneng Zhan, Lingjie Liu, and Christian Theobalt are funded by the ERC Consolidator Grant 4DRepLy (770784).
Yingchen Yu, Rongliang Wu, Jiahui Zhang, and Shijian Lu are supported by the ERC Consolidator Grant 4DRepLy (770784) and the RIE2020 Industry Alignment Fund – Industry Collaboration Projects (IAF-ICP) Funding Initiative.
Adam Kortylewski acknowledges support via his Emmy Noether Research Group funded by the German Science Foundation (DFG) under Grant No. 468670075.

{\small
\bibliographystyle{unsrt2authabbrvpp}
\bibliography{cite}
}

\vspace{-0.75cm}
\begin{IEEEbiographynophoto}{Fangneng Zhan}
is a postdoctoral researcher at Max Planck Institute for Informatics.
He received the Ph.D. degree in Computer Science \& Engineering from Nanyang Technological University. His research interests include generative models and neural rendering.
He serves as a reviewer or program committee member for top journals and conferences including TPAMI, ICLR, ICML, NeurIPS, CVPR, ICCV.
\end{IEEEbiographynophoto}

\vspace{-1cm}
\begin{IEEEbiographynophoto}{Yingchen Yu}
is currently pursuing the Ph.D. degree at School of Computer Science and Engineering, Nanyang Technological University under Alibaba Talent Programme. His research interests are image synthesis and manipulation.
\end{IEEEbiographynophoto}

\vspace{-1cm}
\begin{IEEEbiographynophoto}{Rongliang Wu}
received the Ph.D. degree from School of Computer Science and Engineering, Nanyang Technological University. His research interests include computer vision and deep learning, specifically for facial expression analysis and generation.
\end{IEEEbiographynophoto}

\vspace{-1cm}
\begin{IEEEbiographynophoto}{Jiahui Zhang} 
is currently pursuing the Ph.D. degree at School of Computer Science and Engineering, Nanyang Technological University. His research interests include computer vision and machine learning.
\end{IEEEbiographynophoto}

\vspace{-1cm}
\begin{IEEEbiographynophoto}{Shijian Lu} is an Associate Professor in the School of Computer Science and Engineering, Nanyang Technological University. He received his PhD in Electrical and Computer Engineering from the National University of Singapore. His research interests include computer vision and deep learning. 
He has published more than 100 internationally refereed journal and conference papers. 
Dr Lu is currently an Associate Editor for the journals of Pattern Recognition and Neurocomputing.
\end{IEEEbiographynophoto}

\vspace{-1cm}
\begin{IEEEbiographynophoto}{Lingjie Liu} is the Aravind K. Joshi Assistant Professor in the Department of Computer and Information Science at the University of Pennsylvania. Before that, she was a Lise Meitner postdoctoral researcher in the Visual Computing and AI Department at Max Planck Institute for Informatics. She obtained her Ph.D. degree from the University of Hong Kong in 2019. Her research interests are Neural Scene Representations, Neural Rendering, Human Performance Modeling and Capture, and 3D Reconstruction.
\end{IEEEbiographynophoto}

\vspace{-1cm}
\begin{IEEEbiographynophoto}{Adam Kortylewski} is a research group leader at the University of Freiburg and the Max Planck Institute for Informatics where he leads the Generative Vision and Robust Learning lab. Before that he was a postdoc at Johns Hopkins University with Alan Yuille for three years. He obtained his PhD from the University of Basel with Thomas Vetter. His research focuses understanding the principles that enable artificial intelligence systems to reliably perceive our world through images.  Adam was awarded the prestigious Emmy Noether Grant (2022) of the German Science Foundation for exceptionally qualified early career researchers.
\end{IEEEbiographynophoto}

\vspace{-1cm}
\begin{IEEEbiographynophoto}{Christian Theobalt} is a Professor of Computer Science and the director of the department “Visual Computing and Artificial Intelligence” at the Max Planck Institute for Informatics, Germany. He is also a professor at Saarland University. His research lies on the boundary between Computer Vision and Computer Graphics. Christian received several awards, for instance the Otto Hahn Medal of the Max Planck Society (2007), the EUROGRAPHICS Young Researcher Award
(2009), the German Pattern Recognition Award (2012), an ERC Starting Grant (2013), an ERC Consolidator Grant (2017), and the Eurographics Outstanding Technical Contributions Award (2020). In 2015, he was elected one of Germany’s top 40 innovators under 40 by the magazine Capital.
\end{IEEEbiographynophoto}

\vspace{-1cm}
\begin{IEEEbiographynophoto}{Eric Xing} (Fellow, IEEE) received 
the Ph.D. degree in computer science from the University of California at Berkeley, Berkeley, CA, USA, in 2004. He is currently a Professor of machine learning with the School of Computer Science, Carnegie Mellon University, Pittsburgh, PA, USA. His principal research interests lie in the development of machine learning and statistical methodology, especially for solving problems involving automated learning, reasoning, and decision-making in high-dimensional, multimodal, and dynamic possible worlds in social and biological systems. Dr. Xing is a member of the DARPA Information Science and Technology (ISAT) Advisory Group and the Program Chair of the International Conference on Machine Learning (ICML) 2014. He is also an Associate Editor of The Annals of Applied Statistics (AOAS), the Journal of American Statistical Association (JASA), the IEEE Transactions on Pattern Analysis and Machine Intelligence (T-PAMI), and PLOS Computational Biology and an Action Editor of the Machine Learning Journal (MLJ) and the Journal of Machine Learning Research (JMLR).

\end{IEEEbiographynophoto}

\end{document}